\def\year{2022}\relax
\documentclass[letterpaper]{article} 
\usepackage{aaai22}  
\usepackage{times}  
\usepackage{helvet}  
\usepackage{courier}  
\usepackage[hyphens]{url}  
\usepackage{graphicx} 
\usepackage{natbib}  
\usepackage{caption} 
\DeclareCaptionStyle{ruled}{labelfont=normalfont,labelsep=colon,strut=off} 
\usepackage{algorithm}
\usepackage{algorithmic}

\usepackage{newfloat}
\usepackage{listings}
\floatstyle{ruled}
\newfloat{listing}{tb}{lst}{}
\floatname{listing}{Listing}

\usepackage{amsmath,amssymb,amsthm}

\usepackage{subcaption}

\theoremstyle{definition}
\newtheorem{definition}{Definition}

\newtheorem{proposition}[definition]{Proposition}
\newtheorem{definition/proposition}[definition]{Definition/Proposition}

\nocopyright
\title{Gradual (In)Compatibility of Fairness Criteria}
\author{
    Corinna Hertweck,\equalcontrib\textsuperscript{\rm 1, 2}
    Tim Räz\equalcontrib\textsuperscript{\rm 3}
}
\affiliations{

    \textsuperscript{\rm 1}Institute for Data Analysis and Process Design, Zurich University of Applied Sciences, Winterthur, Switzerland\\
    \textsuperscript{\rm 2}Department of Informatics, University of Zurich, Zurich, Switzerland\\
    \textsuperscript{\rm 3}Institute of Philosophy, University of Bern, Bern, Switzerland\\
    corinna.hertweck@zhaw.ch, tim.raez@posteo.de
}

\begin{document}

\maketitle

\begin{abstract}
Impossibility results show that important fairness measures (independence, separation, sufficiency) cannot be satisfied at the same time under reasonable assumptions. This paper explores whether we can satisfy and/or improve these fairness measures simultaneously to a certain degree. We introduce information-theoretic formulations of the fairness measures and define degrees of fairness based on these formulations. The information-theoretic formulations suggest unexplored theoretical relations between the three fairness measures. In the experimental part, we use the information-theoretic expressions as regularizers to obtain fairness-regularized predictors for three standard datasets. Our experiments show that a) fairness regularization directly increases fairness measures, in line with existing work, and b) some fairness regularizations indirectly increase other fairness measures, as suggested by our theoretical findings. This establishes that it is possible to increase the degree to which some fairness measures are satisfied at the same time -- some fairness measures are gradually compatible.
\end{abstract}

\section{Introduction}

Since the publication of \citet{angwi2016}, statistical fairness measures such as independence (also known as statistical parity and demographic parity), separation and sufficiency have been much discussed. There has been interest in the relation between these fairness measures (see, e.g., \citet{Klein2016, choul2017, mitchell2021algorithmic, heidari2019moral, leben2020normative, baer2019fairness}), and also in the relation between fairness measures and accuracy (see, e.g., \citet{corbett2017algorithmic, menon2018cost, feder2021emergent}).

It is well known that strict satisfaction of some combinations of these fairness measures is not possible; impossibility theorems establish that satisfaction of one fairness measure may prevent satisfaction of another under certain conditions \cite{Klein2016, choul2017, baroc2019}. This is a fundamental limitation if one wishes to satisfy different fairness measures at the same time. However, it remains an open question to what extent fairness criteria are (in)compatible if they are \emph{partially fulfilled}. In this paper, we take first steps towards answering this question. This is relevant in practice because the application of predictive models typically involves multiple stakeholders. These stakeholders might reasonably disagree about the most appropriate fairness metric. It would therefore be helpful to know whether two fairness metrics can be improved at the same time in order to trade off the interests of multiple stakeholders.

In the present paper, we will show that it is possible to simultaneously improve the degree to which some of these fairness measures are satisfied. To show this, we first define a notion of partial satisfaction of fairness measures via information theory. The fairness measures have an exact information-theoretic counterpart, but the information-theoretic formulation also generalizes the fairness measures in providing degrees of fairness. An information-theoretic decomposition of separation and sufficiency reveals quantitative relations between the three measures that have not been explored so far: The expression of separation can be viewed as interpolating between sufficiency and independence. These theoretical relations suggest that enforcing one of the measures may quantitatively improve others.

We then put these theoretical considerations to work using so-called fairness regularization. The information-theoretic formulations of independence, separation and sufficiency serve as regularization terms during training to create fairness-regularized predictors for three standard datasets. We first examine whether fairness regularization directly increases the respective fairness measure after training, showing that this works reasonably well. We then explore whether enforcing one fairness measure indirectly increases the degree to which other fairness measures hold, in keeping with our theoretical hypotheses. We show that enforcing independence indirectly increases separation, and vice versa. We also discover that enforcing sufficiency does not increase separation, and vice versa. We offer a diagnosis of why this is the case on the basis of our theoretical findings.

It is also known that increasing fairness may depreciate accuracy \cite{corbett2017algorithmic}, and that sufficiency and separation may be better aligned with accuracy than independence --- this is true at least for perfect accuracy \cite{hardt2016}. For predictors that are not perfectly accurate, it has been shown that increases of accuracy and fairness need not go hand in hand \cite{raez2021group}. It would be desirable to better understand how accuracy and fairness are related when neither one is perfectly satisfied. The information-theoretic decompositions of separation and sufficiency suggest that accuracy is a natural ingredient of these fairness measures.

The paper is structured as follows: In Section \ref{sec:theory}, we introduce the information-theoretic formulations of statistical fairness measures and decompose them in order to examine how the fairness measures are related. We also formulate empirical hypotheses about the effect of using these information-theoretic formulations in fairness regularization. Section \ref{sec:contextualization} situates our contribution in the broader recent discussion of fair-ML while Section \ref{sec:background} discusses the relation of our contribution to similar works. In Section \ref{sec:experiments}, we experimentally test our previously stated hypotheses. For this, we provide descriptions of the experimental setup and discuss our empirical results on direct and indirect fairness regularization. In Section \ref{sec:discussion}, we summarize our findings and highlight open questions as well as avenues for future research.

\section{Theory and Hypothesis Formulation}\label{sec:theory}

\subsection{Preliminaries}

We first provide some standard definitions from information theory; see \citet{cover2006} for background. All random variables are discrete. Two random variables $X,Y$ are (statistically) \emph{independent}, written as  $X \perp Y$, if $P(X,Y) = P(X)\,P(Y)$; two random variables $X, Y$ are \emph{conditionally independent} given $Z$, written as $X\perp Y \mid Z$, if $P(X\mid Y, Z) = P(X\mid Z)$. The \emph{entropy} of a random variable $X \sim p(x)$ is given by

\begin{equation*}
H(X) = - \sum_x p(x)\, log\, p(x); 
\end{equation*}

the \emph{joint entropy} of a pair of random variables $(X,Y) \sim p(x,y)$ is given by

\begin{equation*}
H(X,Y) = - \sum_{x, y} p(x,y)\, log\, p(x,y);
\end{equation*}

the \emph{conditional entropy} of a pair of random variables $(X,Y) \sim p(x,y)$ is given by

\begin{equation*}
H(Y|X) = - \sum_{x, y} p(x,y)\, log\, p(y|x).
\end{equation*}

The \emph{mutual information} of a pair of random variables $(X,Y) \sim p(x,y)$ is given by

\begin{equation*}
I(X;Y) = H(X) - H(X|Y);
\end{equation*}

the \emph{conditional mutual information} of three random variables $(X,Y,Z) \sim p(x,y,z)$ is given by

\begin{equation*}
I(X;Y\mid Z) =  H(X\mid Z) - H(X|Y,Z).
\end{equation*}

\subsection{Information-Theoretic Fairness}

We can now define statistical notions of fairness (following \citet{baroc2019}) in probabilistic and information-theoretic terms. We work with discrete random variables $Y$, $R$, $A$, where $Y$ is the ``ground truth'', $R$ is the prediction, and $A$ is the sensitive attribute.

\begin{definition}
The \emph{independence gap} is given by
\begin{equation}
I(R;A).
\tag{IND}
\label{independence}
\end{equation}
Independence is satisfied if $R\perp A$, which is the case iff. $I(R;A) = 0$. We define the independence gap to be of degree $d$ if $I(R;A) \leq d$.
\end{definition}

\begin{definition}
The \emph{separation gap} is given by
\begin{equation}
I(R;A \mid Y).
\tag{SEP}
\label{separation}
\end{equation}
Separation is satisfied if $R \perp A \mid Y$, which is the case iff. $I(R;A \mid Y) = 0$. The separation gap is of degree $d$ if $I(R;A|Y) \leq d$.
\end{definition}

\begin{definition}
The \emph{sufficiency gap} is given by
\begin{equation}
I(Y;A\mid R).
\tag{SUF}
\label{sufficiency}
\end{equation}
Sufficiency is satisfied if $Y \perp A \mid R$, which is the case iff. $I(Y;A\mid R) = 0$. The sufficiency gap is of degree $d$ if $I(Y;A|R) \leq d$.
\end{definition}

\begin{definition}
The \emph{accuracy} of a predictor $R$ is given by
\begin{equation}
I(Y;R).
\tag{ACC}
\end{equation}
\end{definition}

Note that while fairness measures are enforced by minimizing the corresponding expression, accuracy is enforced by maximizing the corresponding expression. In what follows, we will refer to the information-theoretic expressions of fairness metric gaps as independence, separation and sufficiency to improve readability.

The following definition is motivated in the next section.

\begin{definition}
\emph{Balance} is given by
\begin{equation}
I(Y;R \mid A).
\tag{BAL}
\end{equation}
\end{definition}

\subsection{Decomposition of Sufficiency and Separation}

In this section, we use the information-theoretic formulations of sufficiency and separation given above to gain a clearer picture of how these notions are related to each other, as well as to independence and accuracy.

\begin{proposition}
Sufficiency has the following decomposition:
\begin{equation}
I(Y;A\mid R) =  -\underbrace{I (Y;R)}_{accuracy} + \underbrace{I(Y;R \mid A)}_{balance} + \underbrace{I(A;Y)}_{legacy}.
\label{suff decomposition}
\end{equation}
\end{proposition}

The proof of the proposition can be found in Appendix \ref{app:proofs}, which is available in the extended version of this paper.\footnote{\url{https://arxiv.org/abs/2109.04399}} The first term on the right of equation (\ref{suff decomposition}) corresponds to accuracy, while we refer to the second term as balance. If accuracy increases, $R$ contains more information about $Y$ and the first term becomes smaller due to the negative sign. If $Y$ and $R$ become decorrelated given $A$, the second term becomes smaller. The third term is dubbed \emph{legacy} term because it only depends on the distribution of $(Y,A)$, i.e., on how we find the world to be at a certain point, measured by the ``true label'' $Y$ and the characteristic $A$, but not on the prediction $R$ (see \citet{kamis2011, raez2021group} for more on this). Only the first and the second term depend on the prediction $R$; changing $R$ will not affect the legacy term.\footnote{The first two terms on the right are also known as \emph{interaction information}; see, e.g., \citet{willi2010}. Interaction information measures the extent to which the interaction between two variables (here $Y, R$) is mediated by a third one (here $A$). Interaction information can be negative, zero, and positive. If we can only control $R$, but not $Y$ and $A$, then changes in interaction information determine changes in sufficiency.}

\begin{proposition}
Separation has the following decomposition:
\begin{equation}
I(R;A \mid Y) = -\underbrace{I(Y;R)}_{accuracy} + \underbrace{I(Y;R\mid A) }_{balance} + \underbrace{I(A;R)}_{independence}.
\label{sep decomposition}
\end{equation}
\end{proposition}

The proof of the proposition is analogous to the proof of the proposition above, see also Appendix \ref{app:proofs}. In the decomposition of separation, three terms depend on $R$. The first two terms on the right of equation (\ref{sep decomposition}) are the same as in sufficiency; the third term corresponds to independence. The fact that all three terms depend on $R$ means that if we want to see how changing $R$ affects separation, we need to take all three terms into account. This is a major difference between separation and sufficiency.

\subsection{Empirical Hypotheses}

In this section, we collect some observations about the relations between the three fairness measures that can be gleaned from the decompositions in the previous section. This will allow us to formulate empirical hypotheses.

A first observation concerns the fact that both sufficiency and separation contain the accuracy term with a negative sign. This is related to the fact that sufficiency and separation are both what \citet{raez2021group} refers to as \emph{conservative} measures, i.e., they are both minimal in case of perfect prediction ($Y=R$). However, as has been shown in \citet{raez2021group}, we do not necessarily improve sufficiency and separation by increasing accuracy. The decompositions above show why this is the case: If the accuracy of the predictor $R$ is increased, this will only lower sufficiency and separation respectively if the other summands containing $R$ do not offset the change.\footnote{A predictor $R'$, which is at least as accurate as predictor $R$, maintains the degree of sufficiency of $R$ if $I(Y;R' \mid A) \leq I(Y;R \mid A)$, and the predictor $R'$ maintains the degree of separation of $R$ if $I(Y;R' \mid A) + I(A;R') \leq I(Y;R \mid A) + I(A;R)$.} Note that while accuracy, as it is defined here, is not commonly used as a loss function in machine learning, it can be expected to be closely correlated with common loss functions such as cross-entropy. This suggests that it is possible to indirectly increase the degree to which both sufficiency and separation are satisfied by making a predictor more accurate, \emph{given} that one controls for the other terms containing $R$ in the above decompositions. This would be in line with work by \citet{liu2019implicit} who found that group calibration, a form of sufficiency, increases with increased accuracy. As models are typically optimized for some variation of accuracy, they note that sufficiency is often already achieved by unconstrained learning, i.e., when no fairness constraints are imposed.

A second observation also concerns the relation between separation and sufficiency. Apart from being related to accuracy, we can observe that they both also contain the balance term. This suggests that one way of improving both measures would be to control or decrease the balance term.

A third observation concerns the relation between all three measures. Equation \ref{sep decomposition} shows that independence is one of the summands of the decomposition of separation. On the other hand, independence does not contribute to sufficiency. In this sense, we can think of separation as being an intermediate between independence and sufficiency, or as being an interpolation between independence and sufficiency.

These observations lead to the following idea. The various relations between the three measures just pointed out mean that the three measures co-vary to a certain extent, in that if, say, independence is higher, then separation may also be higher. These possible co-variations may be exploited when fairness measures are enforced: If we increase the degree to which a predictor satisfies one of the measures, this may indirectly increase or decrease one of the other measures.

In Section \ref{sec:experiments} we will test this general idea in the form of direct and indirect fairness regularization. Fairness regularization means that the information-theoretic formulations of the fairness measures can be used as a regularization constraint in training an ML model --- \citet{kamis2011} call this \emph{prejudice remover regularization}. Indirect fairness regularization means that the relations between the fairness measures may lead to an increase of a fairness measure even if it is not directly regularized, due to the relations between the measures identified above.

First, we will test for \emph{direct regularization}:

\begin{itemize}

\item \emph{Independence $\rightarrow$ Independence, Separation $\rightarrow$ Separation, Sufficiency $\rightarrow$ Sufficiency:} If we use the three information-theoretic expressions of independence, separation and sufficiency as regularization terms during training, this should increase the degree to which they are satisfied.

\end{itemize}

Second, we will test for \emph{indirect regularization:}

\begin{itemize}

\item \emph{Independence $\rightarrow$ Separation:} If we regularize independence, then this could indirectly decrease separation, due to the fact that the independence term features in the decomposition of separation in equation \ref{sep decomposition}.

\item \emph{Separation $\rightarrow$ Independence:} Regularizing separation could indirectly regularize independence because independence is in the decomposition of separation.

\item \emph{Separation $\rightarrow$ Sufficiency:} Balance and accuracy are in the decomposition of both sufficiency and separation, so regularizing separation could indirectly regularize sufficiency.

\item \emph{Sufficiency $\rightarrow$ Separation:} Balance and accuracy are in the decomposition of both sufficiency and separation, so similarly, regularizing sufficiency could indirectly regularize separation.

\item \emph{Balance $\rightarrow$ Separation, Sufficiency:} Balance is in the decomposition of both sufficiency and separation, so regularizing balance could indirectly regularize these two measures.

\item \emph{Accuracy $\rightarrow$ Separation, Sufficiency:} Accuracy is in the decomposition of both sufficiency and separation, so regularizing accuracy could indirectly regularize these two measures.

\end{itemize}

\subsection{Normalizations}

It is desirable to compare normalized versions of the fairness measures to get a better sense of how the fairness measures perform during evaluation. We therefore normalize independence, separation and sufficiency in the evaluation step (but not during training). The normalizations we use here were first proposed and justified in \cite{steinberg2020fast}. The normalization for \emph{independence} is

\begin{equation*}
\hat{I}(R;A) = \frac{I(R;A)}{H(A)}. 
\tag{N-IND}
\end{equation*}

The justification given by Steinberg et al. is that in case $\hat{I}(R;A) = 1$, which is the maximum, $R$ encodes all information about $A$, which is maximally unfair. The justifications for the normalizations of separation and sufficiency are similar. The normalization for \emph{separation} is

\begin{equation*}
\hat{I}(R;A|Y) = \frac{I(R;A|Y)}{H(A|Y)}.
\tag{N-SEP}
\end{equation*}

The normalization for \emph{sufficiency} is

\begin{equation*}
\hat{I}(Y;A|R) = \frac{I(Y;A|R)}{H(A|R)}.
\tag{N-SUF}
\end{equation*}

The normalization factors are (conditional) entropies. The entropies for independence and separation do not depend on $R$ and have to be computed only once per dataset. The entropy for sufficiency involves $R$ and has to be computed for each (trained) model, i.e., for every evaluation.

The normalized values for independence, separation and sufficiency thus vary between 0 and 1, where 0 means that the fairness metric is perfectly fulfilled, and 1 indicates maximal unfairness. It is an open question what (philosophical) meaning can be ascribed to values between these two extremes. What is known is that lower values indicate a better fulfillment of the respective fairness metric.

\section{Background}\label{sec:contextualization}

In 2016, ProPublica published an article entitled "Machine Bias," in which they investigated the tool COMPAS (short for "Correctional Offender Management Profiling for Alternative Sanctions") \cite{angwi2016}. ProPublica accused the company that developed the software, Northpointe (now Equivant), of racial bias in their tool. According to their calculations, black defendants who are not arrested within two years are deemed more likely to be arrested than white defendants who are arrested within two years. By comparing risk predictions ($R$) conditioned on rearrest ($Y$) across groups, they argued that a version of separation is violated. Northpointe denied the accusation of racial bias by arguing that their risk scores are equally accurate for black and white defendants \cite{diete2016}. They thus argued that a version of sufficiency was fulfilled. For further contributions to this discussion see \citet{flore2016, berk2018}. Subsequently, \citet{Klein2016} and \citet{choul2017} demonstrated that separation has to be violated when sufficiency is fulfilled and vice versa -- except under circumstances that cannot be expected to arise in practice. This incompatibility debate has been extended in \citet{baroc2019}: Any two of the three fairness criteria, i.e., independence, separation and sufficiency, are mutually incompatible under reasonable assumptions and at least for a binary $Y$.

Our paper investigates whether mutually exclusive fairness metrics can still be jointly improved to a certain extent. In doing so, we use an approach that \citet{green2021impossibility} calls "formal": We seek to improve predictive models and do not consider possible changes in the social context of their application. Green criticizes that while the "formalism response" can sometimes increase social justice, its narrow focus on mathematical formalizations of egalitarian principles can also be harmful because the social context of the application is not considered. As pointed out by \citet{green2018myth} and \citet{bao2021its}, incremental improvements in the fulfillment of mathematical definitions can distract from real reforms that target unjust systems (over predictive models that are just a small part of these systems).

We agree that the social context in which models are deployed must not be ignored when judging the fairness of models. However, in the present paper, we merely investigate the relations between three mathematical definitions of fairness. Our results are proofs-of-concept for the partial fulfillment of multiple fairness metrics. The possibility of improving fairness metrics does \emph{not} establish that fairness has been achieved, even incrementally. If direct reforms of unjust systems are not a viable option, improving the fulfillment of fairness metrics may be necessary. In such cases, understanding the relationship between the fairness metrics and possibilities for tradeoffs between them is important for improving fairness.

In our experiments, we also investigate the effects of fairness regularization on the accuracy of the predictive model. The tradeoff between fairness and accuracy has been analyzed in several works \cite{hardt2016, corbett2017algorithmic, menon2018cost, chen2018my}. However, \citet{feder2021emergent} point out that this framing can potentially cause more unfairness because we usually work with specific operationalizations of both fairness and accuracy, which may be problematic. The goal of recidivism predictions, for example is to predict if an incarcerated person would \textit{commit} a crime after being released. This is typically operationalized as the question of who is \textit{arrested} within a given time period after release. However, it is at least questionable if this operationalization of \textit{crime} as \textit{arrest} works as intended due to unjustified arrests and undetected crimes; see also \citet{lum2016,mehro2021,jacobs2021measurement}. Moreover, accuracy does not take long-term effects of, say, imposing fairness constraints into account. So even though avoiding a loss in accuracy appears desirable from a mathematical perspective, we have to keep in mind that accuracy, as it is typically measured in ML, does not necessarily represent real-world accuracy.

\section{Related Work}\label{sec:background}

\citet{kamis2011} proposed mitigating bias in predictive models by using a version of independence for regularization. For this, they formalized independence in information-theoretic terms and simplified the expression further to make it computationally tractable. They used the resulting simplified expression as a regularizer during training, and they found that this improved their definition of fairness, which is closely related to independence, but compromised on accuracy. 
\citet{steinberg2020fast} proposed approximations of information-theoretic terms of independence, separation and sufficiency for mixed variables that they find to be faster to compute than the exact terms, but lead to similar results when used as regularizers. They considered decompositions of fairness measures similar to our equations (\ref{suff decomposition}) and (\ref{sep decomposition}) above, but they did not further investigate the relations between fairness measures. 
\citet{ghassami2018fairness} propose a method for maximizing accuracy while fulfilling separation using an information-theoretic version of it. The described approaches focus on fairness-accuracy-tradeoffs; our focus, on the other hand, is wider because we consider accuracy as only one component of fairness measures as found in the fairness decompositions.

Fairness regularization of some fairness measures using non-information-theoretic regularization terms have also been investigated, e.g., in \citet{cho2020}, \citet{naras2018}, \citet{padal2020} and \citet{chuang2021fair}. These works focus on improving fairness measures through regularization, but they do not investigate how the regularization of one measure affects other measures through indirect regularization, as far as we can see.

Partial fulfillment of fairness metrics has previously been investigated by \citet{pleiss2017fairness} who showed that sufficiency is compatible with a relaxation of separation, i.e., parity in the false negative rates. \citet{loi2019philosophical} discuss conditions for the enforcement of separation and sufficiency. They point out that "when neither set of conditions obtains, fair equality of chances is satisfied by neither sufficiency nor separation, but, possibly, some other statistical constraint" \cite[p. 25]{loi2019philosophical}. This other statistical constraint might be found in the form of a tradeoff between the fairness metrics. Note, however, that even if it can be shown that some conditions can be partially fulfilled or improved simultaneously, it is still necessary to address the philosophical questions as to what partial fulfillment of a fairness metrics means, and whether this is desirable.
\section{Experiments}\label{sec:experiments}

The code for the experimental part can be found in the public repository \url{https://github.com/hcorinna/gradual-compatibility}.

\subsection{Description of Datasets}

Experiments were performed on three standard benchmark datasets, which are briefly described in Table \ref{tab:datasets}: ``Adult income'', ``German credit'', and ``ProPublica recidivism'' (the latter is also known as the ``COMPAS'' dataset). Friedler et al.'s (\citeyear{friedler2019comparative}) preprocessed versions of these datasets were used.
Specifically, we used the preprocessed versions in which categorical values were one-hot encoded and in which sensitive attributes were transformed to binary variables.\footnote{These are the datasets that have the ending ``numerical-binsensitive'' in \citeauthor{friedler2019comparative}'s repository: \url{https://github.com/algofairness/fairness-comparison/tree/master/fairness/data/preprocessed}.}
In addition to Friedler et al.'s (\citeyear{friedler2019comparative}) preprocessing, we preprocess the ProPublica recidivism dataset ourselves by removing the column ``c\_charge\_desc'' (more specifically, the columns resulting from the one-hot encoding of this column).
The column describes the charges raised against the defendant, but as it can take on many different values, it creates a lot of additional columns when one-hot encoded.
We opted for the removal of these columns because this reduces the number of columns from 403 to 14 and thus allows for far faster training.

\begin{table}[h]
\begin{tabular}{lllr}
\hline
\textbf{Dataset} & \textbf{Y=1}              & \textbf{A}                                     & \textbf{Size} \\ \hline
German         & Creditworthiness              & Sex & 1\,000         \\
ProPublica                & Rearrest within 2 years & Race                     & 6\,167         \\
Adult          & Income \textgreater 50k   & Race                                                             & 30\,162        \\ \hline
\end{tabular}
\caption{Datasets used in experiments.}
\label{tab:datasets}
\end{table}

\subsection{Description of Experiments}

In contexts where fairness is particularly salient, it is essential that the decision-making process can be fully understood \cite{rudin2019stop}. Therefore, we chose logistic regression as an easily interpretable model for our proof-of-concept. We use a logistic regression model that is optimized by \texttt{scipy}'s \cite{scipy} \texttt{minimize} function with L-BFGS-B \cite{byrd1995limited, zhu1997algorithm} as the solver.
The loss function takes the following form:

\begin{equation}
L = l_{fit} + \lambda \cdot l_2 + \mu \cdot l_{fair}.
\end{equation}

The algorithm optimizes the cross-entropy loss (dubbed $l_{fit}$) with $l_2$ regularization and fairness regularization terms (dubbed $l_{fair}$). The parameter $\lambda$ is chosen using 5-fold cross validation, for each dataset and before fairness regularization. The value of $\lambda$ was left unchanged during fairness regularization. We used the following values for $\lambda$: 0.00001 (``Adult income''), 0.001 (``German credit''), and 0.001 (``ProPublica recidivism'').\footnote{Note that $\lambda$ could also be determined by cross-validation for each fold of the fairness regularization separately in order to keep parameter tuning and testing strictly separate. We decided to use one value of $\lambda$ per dataset for all fairness regularization experiments because our main goal is not to optimize for accuracy, but to understand relations between fairness measures.} The fairness regularization term $l_{fair}$ takes values in the range \{IND, SEP, SUF, BAL, -ACC\}. We regularize (and thus try to minimize) \textit{negative accuracy}, which is the same as maximizing accuracy because it is desirable to obtain an accurate predictor, and because negative accuracy features in the fairness decompositions. For each fairness regularization term, we choose values for the fairness parameter $\mu$ in the range $[0, 100]$.\footnote{We experimented with $\mu \in [0, 190]$, but did not observe notable changes for $\mu > 100$, so we restrict the plots to the range $[0, 100]$.} The model is trained and evaluated separately for each $\mu$ and $l_{fair}$ using 5-fold cross validation. In the evaluation step, we evaluate the three normalized fairness measures independence, sufficiency, and separation, as well as the negative accuracy, and the non-normalized balance term.\footnote{We also evaluated cross-entropy, but do not show the results for cross-entropy here because accuracy and cross-entropy are very similar in our experiments.} Normalization was only used during evaluation. More details on the experimental setup can be found in the appendix.

\subsection{Direct Regularization}

\begin{figure*}[ht] 
     \centering
     \begin{subfigure}[b]{0.3\textwidth}
         \centering
         \includegraphics[width=\textwidth]{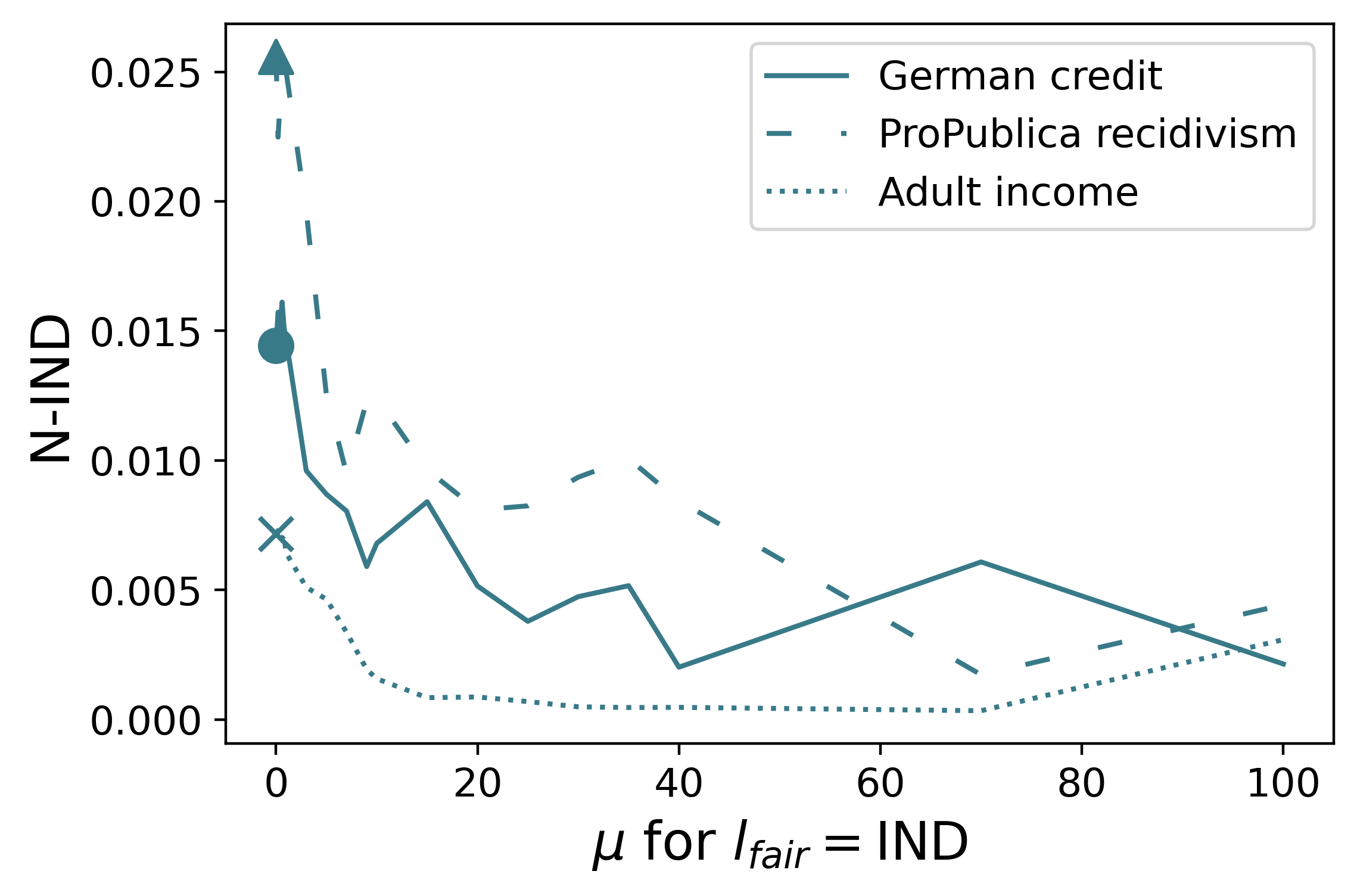}
         \caption{Independence $\rightarrow$ Independence}
         \label{fig:IND--N-IND}
     \end{subfigure}
     \hfill
     \begin{subfigure}[b]{0.3\textwidth}
         \centering
         \includegraphics[width=\textwidth]{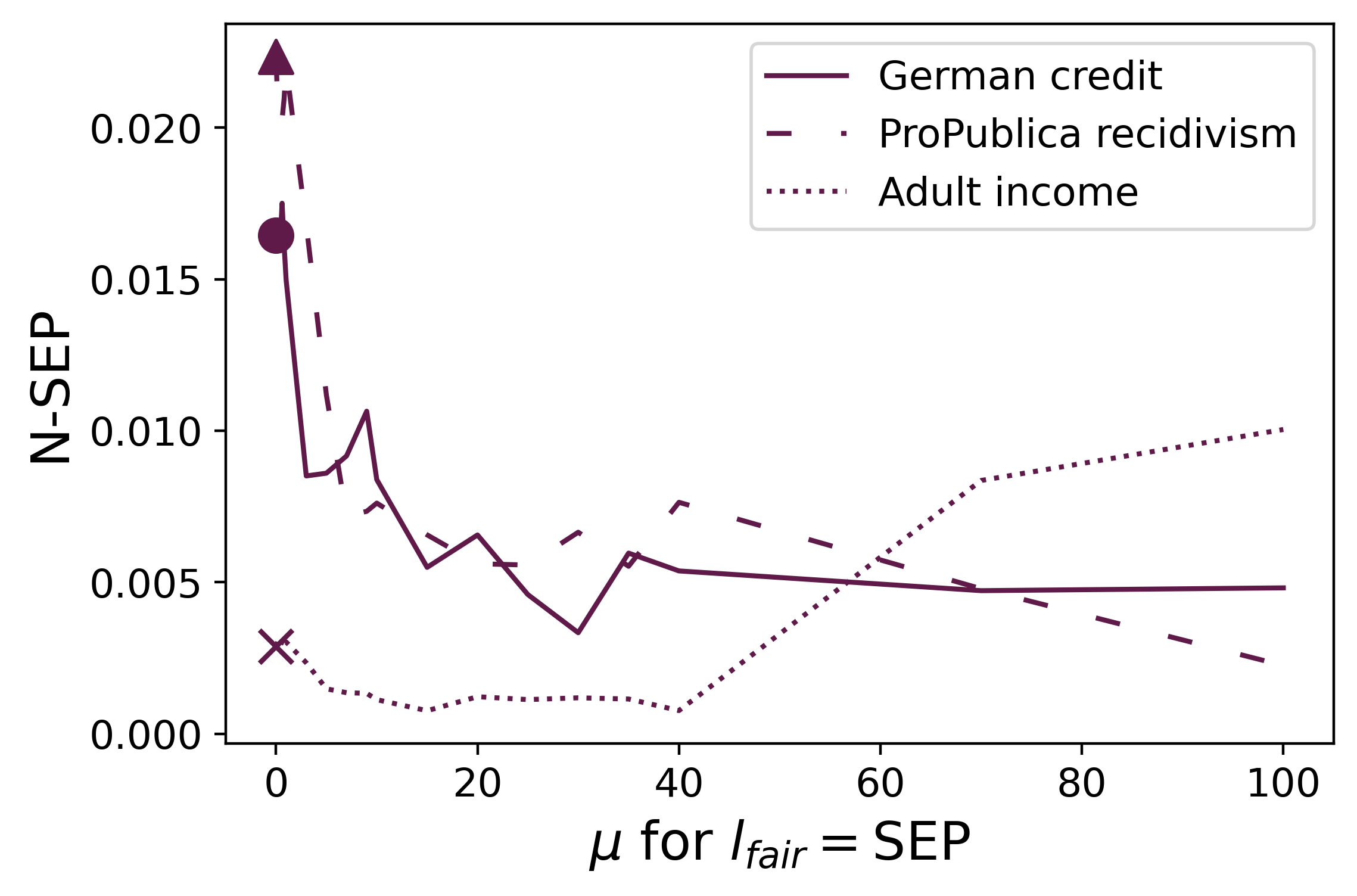}
         \caption{Separation $\rightarrow$ Separation}
         \label{fig:SEP--N-SEP}
     \end{subfigure}
     \hfill
     \begin{subfigure}[b]{0.3\textwidth}
         \centering
         \includegraphics[width=\textwidth]{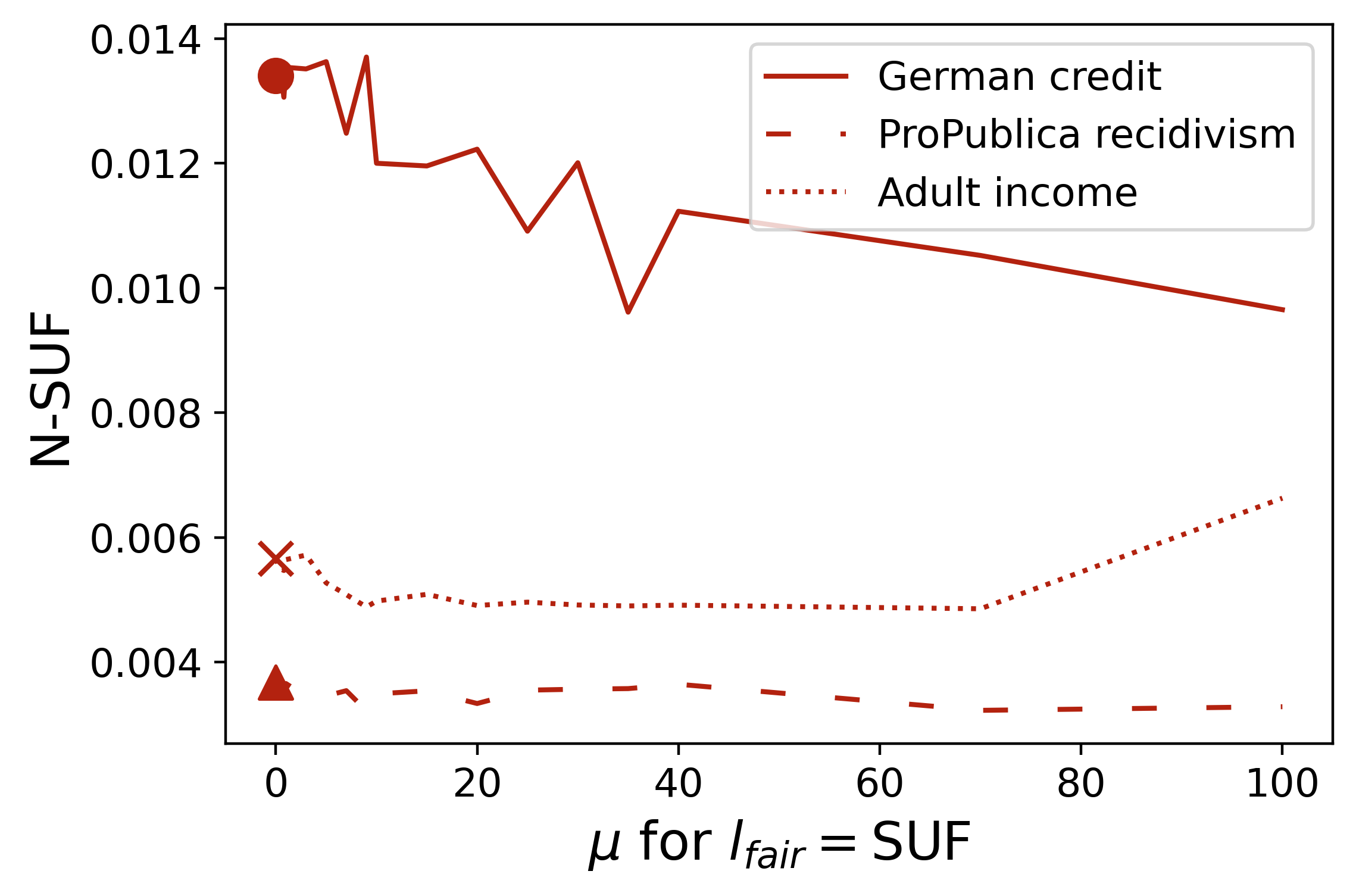}
         \caption{Sufficiency $\rightarrow$ Sufficiency}
         \label{fig:SUF--N-SUF}
     \end{subfigure}
        \caption{Direct regularization. x-axis: Value of $\mu$ in $l_{fair}$. y-axis: Normalized values of regularized metric where lower values are better. The different curves represent the different datasets.}
        \label{fig:direct-regularization}
\end{figure*}

We first evaluate whether the regularization of the three fairness metrics improves the respective (normalized) fairness metric itself, see Figure \ref{fig:direct-regularization}. In Figure \ref{fig:IND--N-IND}, we use independence as the fairness regularizer, that is, we let $l_{fair} = \text{IND}$, train models for different values of $\mu$, and measure N-IND. As expected, independence improves: it decreases in the plot for increasing $\mu$, as more weight is put on the independence regularizer. This is in line with the findings by \citet{kamis2011}. Figure \ref{fig:SEP--N-SEP} shows that regularizing with separation (SEP) improves separation (N-SEP) globally for two datasets, and in a low parameter range for ``Adult income". The regularization of sufficiency (SUF) seems to improve sufficiency (N-SUF) marginally and, for ``Adult income'', only in a low parameter range, see Figure \ref{fig:SUF--N-SUF}. Thus, direct regularization seems to work, but not equally well for all three metrics.

As Figure \ref{fig:effects-on-ACC} in the Appendix shows, the independence and separation regularization have comparable negative effects on accuracy, while sufficiency regularization does not have a pronounced effect on accuracy. We made similar observations for cross-entropy.

We can interpret these results based on the decompositions of sufficiency and separation in equations (\ref{suff decomposition}) and (\ref{sep decomposition}). Regularizing sufficiency does not yield the same improvements as regularizing independence and separation because, first, regularization using the IND term works well and affects independence as well as separation, which contains independence in its decomposition, while independence is not in the decomposition of sufficiency. Second, sufficiency is more closely associated with increasing accuracy, as noted by \citet{liu2019implicit}. However, accuracy is already enforced by the cross-entropy optimization ($l_{fit}$). It is therefore plausible that sufficiency is already naturally close to optimal in unconstrained learning and does not improve as much through fairness regularization.

\subsection{Indirect Regularization}

Now we examine whether the regularization of one fairness metric has effects on other fairness metrics.

\paragraph{Independence and Separation}

\begin{figure}[ht] 
     \centering
     \begin{subfigure}[b]{0.23\textwidth}
         \centering
         \includegraphics[width=\textwidth]{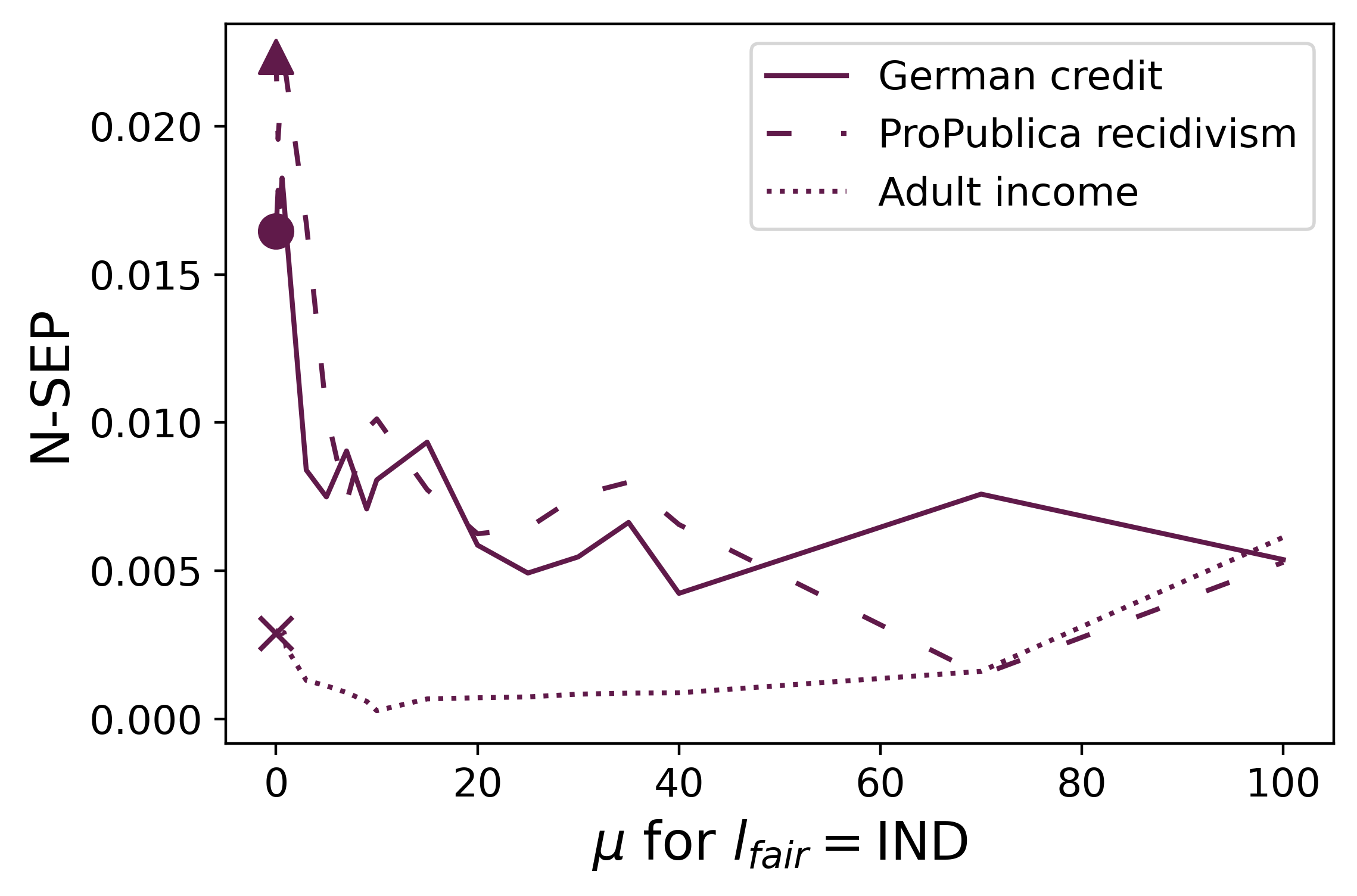}
         \caption{Independence $\rightarrow$ Separation}
         \label{fig:IND--N-SEP}
     \end{subfigure}
     \hfill
     \begin{subfigure}[b]{0.23\textwidth}
         \centering
         \includegraphics[width=\textwidth]{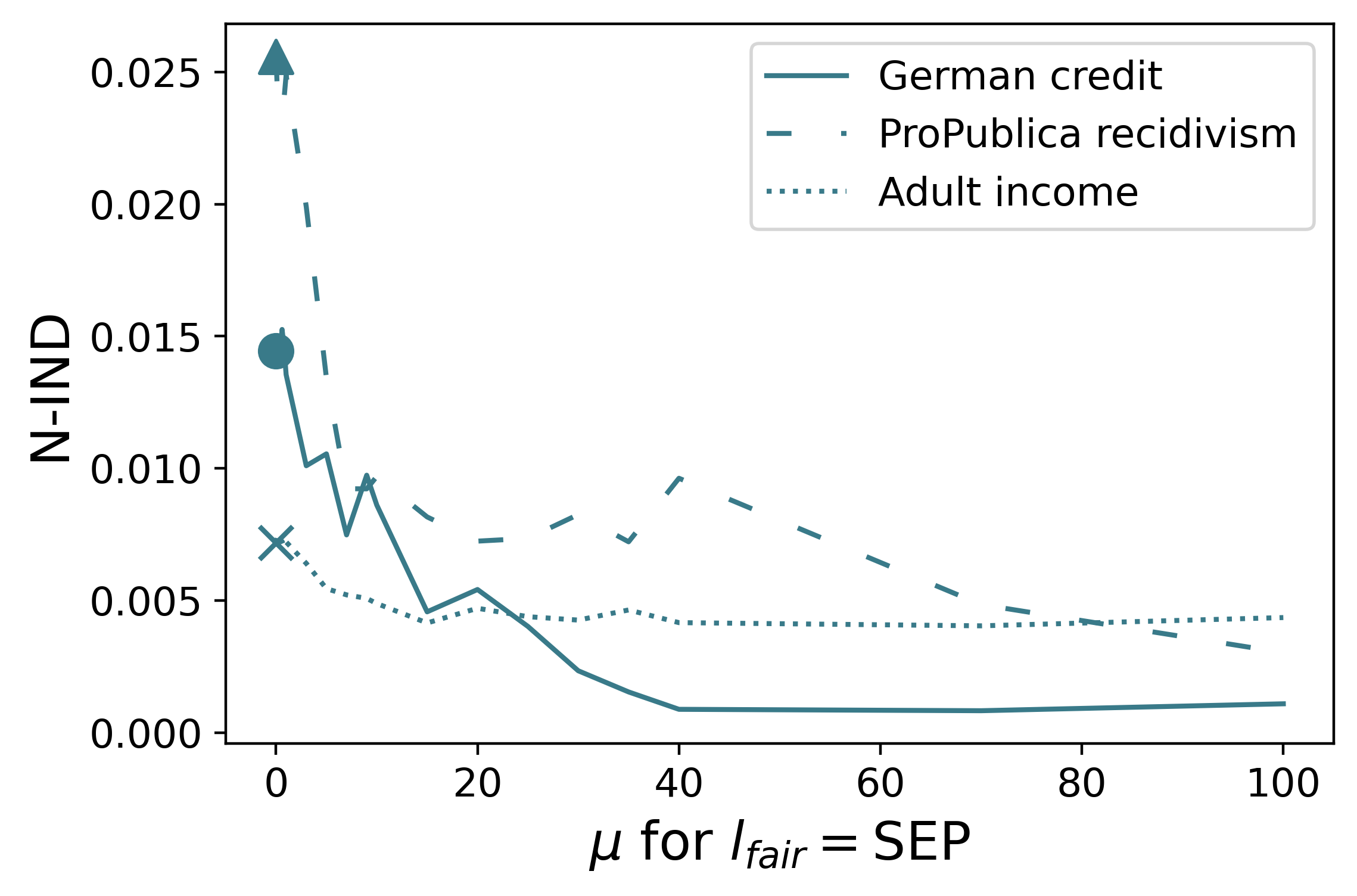}
         \caption{Separation $\rightarrow$ Independence}
         \label{fig:SEP--N-IND}
     \end{subfigure}
     \hfill
     \begin{subfigure}[b]{0.23\textwidth}
         \centering
         \includegraphics[width=\textwidth]{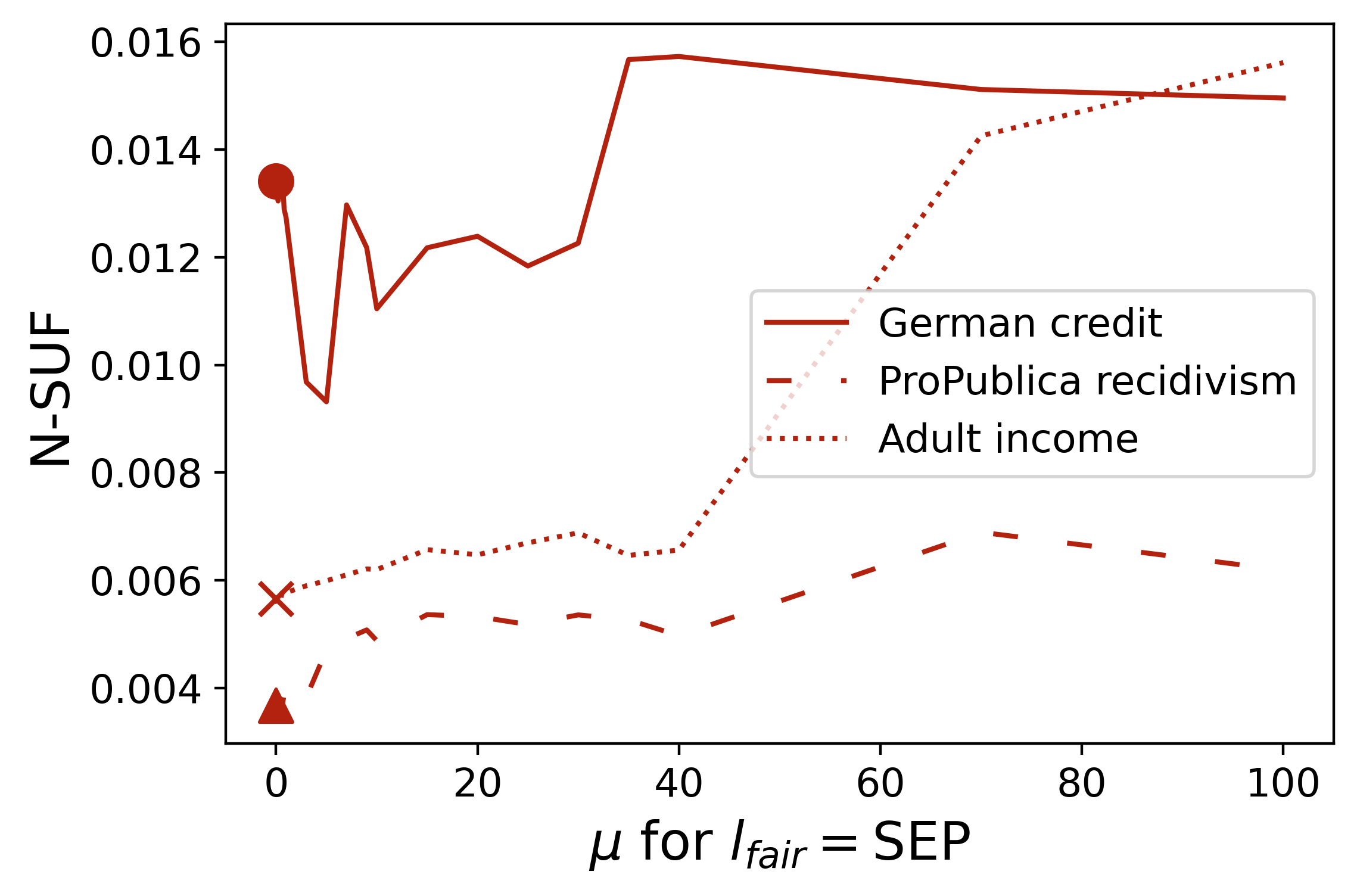}
         \caption{Separation $\rightarrow$ Sufficiency}
         \label{fig:SEP--N-SUF}
     \end{subfigure}
     \hfill
     \begin{subfigure}[b]{0.23\textwidth}
         \centering
         \includegraphics[width=\textwidth]{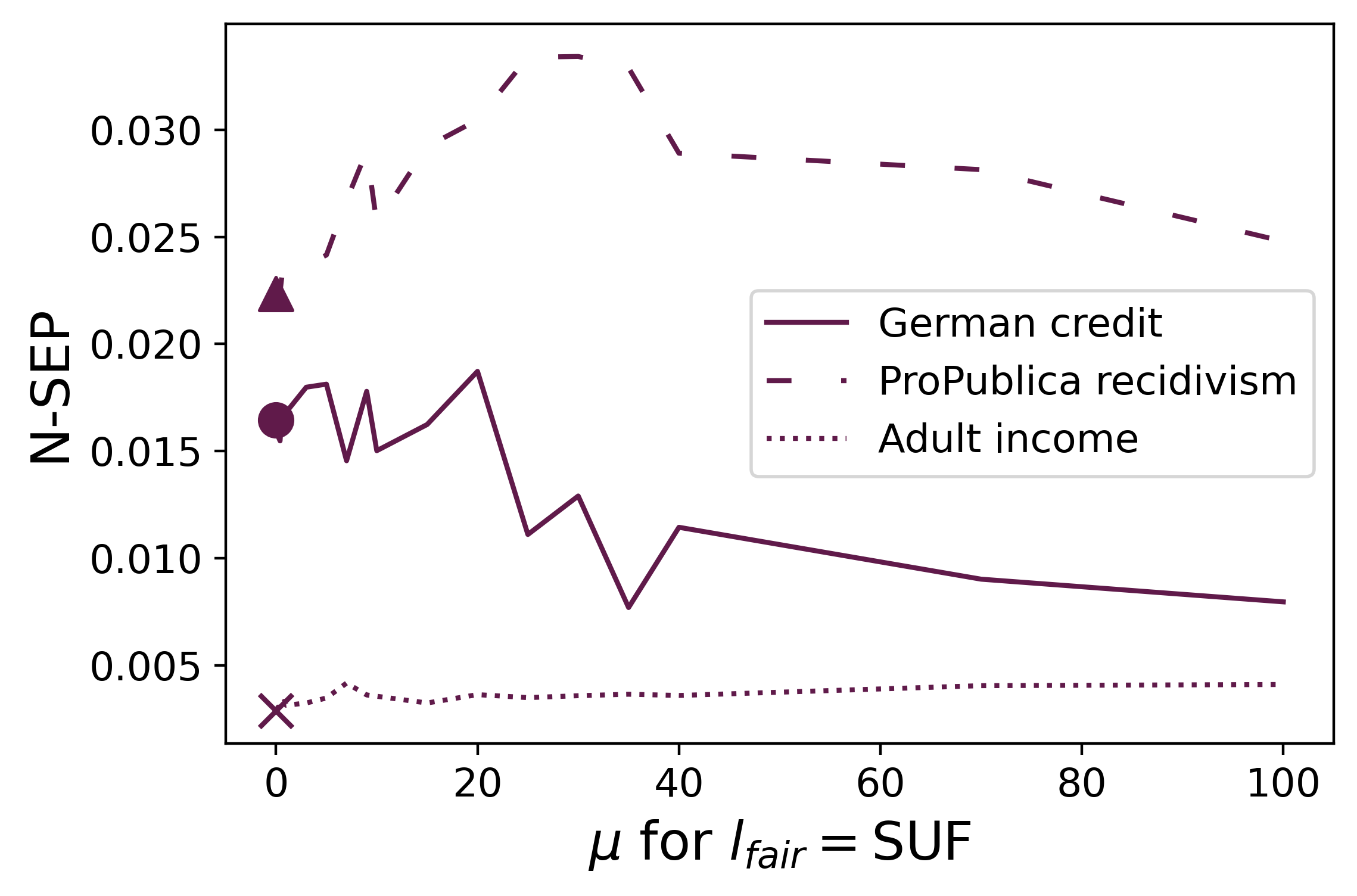}
         \caption{Sufficiency $\rightarrow$ Separation}
         \label{fig:SUF--N-SEP}
     \end{subfigure}
        \caption{Indirect regularization between independence and separation and between separation and sufficiency.}
        \label{fig:indirect-regularization}
\end{figure}

We expect that the regularization of independence and separation have an indirect effect on each other because they share the IND term. Figure \ref{fig:IND--N-SEP} shows the effects of the independence regularizer (IND) on separation (N-SEP). As we can see, separation indeed improves with increasing independence regularization. Figure \ref{fig:SEP--N-IND} shows that the regularization of separation (SEP) has a positive effects on independence (N-IND).

While the impossibility results by \citet{baroc2019} proved that it is impossible to perfectly fulfill independence and separation when ``$Y$ is binary, $A$ is not independent of $Y$, and $R$ is not independent of $Y$," our results show that the two metrics can both be improved simultaneously -- provided they are not perfectly satisfied.

\paragraph{Separation and Sufficiency}

We are hoping for some indirect regularization effects between separation and sufficiency because they share the BAL and -ACC terms. However, we get mixed or negative results, depending on the dataset. In Figure \ref{fig:SEP--N-SUF}, we show the effect of the separation regularizer (SEP) on sufficiency (N-SUF), and in Figure \ref{fig:SUF--N-SEP}, we plot the effect of sufficiency (SUF) regularization on separation (N-SEP). The only positive outcome is that regularizing sufficiency seems to improve separation for the ``German credit'' dataset. 

Indirect regularization does not achieve the desired effect of improving \textit{both} separation and sufficiency at the same time. From this, we conclude that it is difficult to circumvent the impossibility results concerning separation and sufficiency.

\paragraph{Accuracy and Balance}

\begin{figure}[ht] 
     \centering
     \begin{subfigure}[b]{0.23\textwidth}
         \centering
         \includegraphics[width=\textwidth]{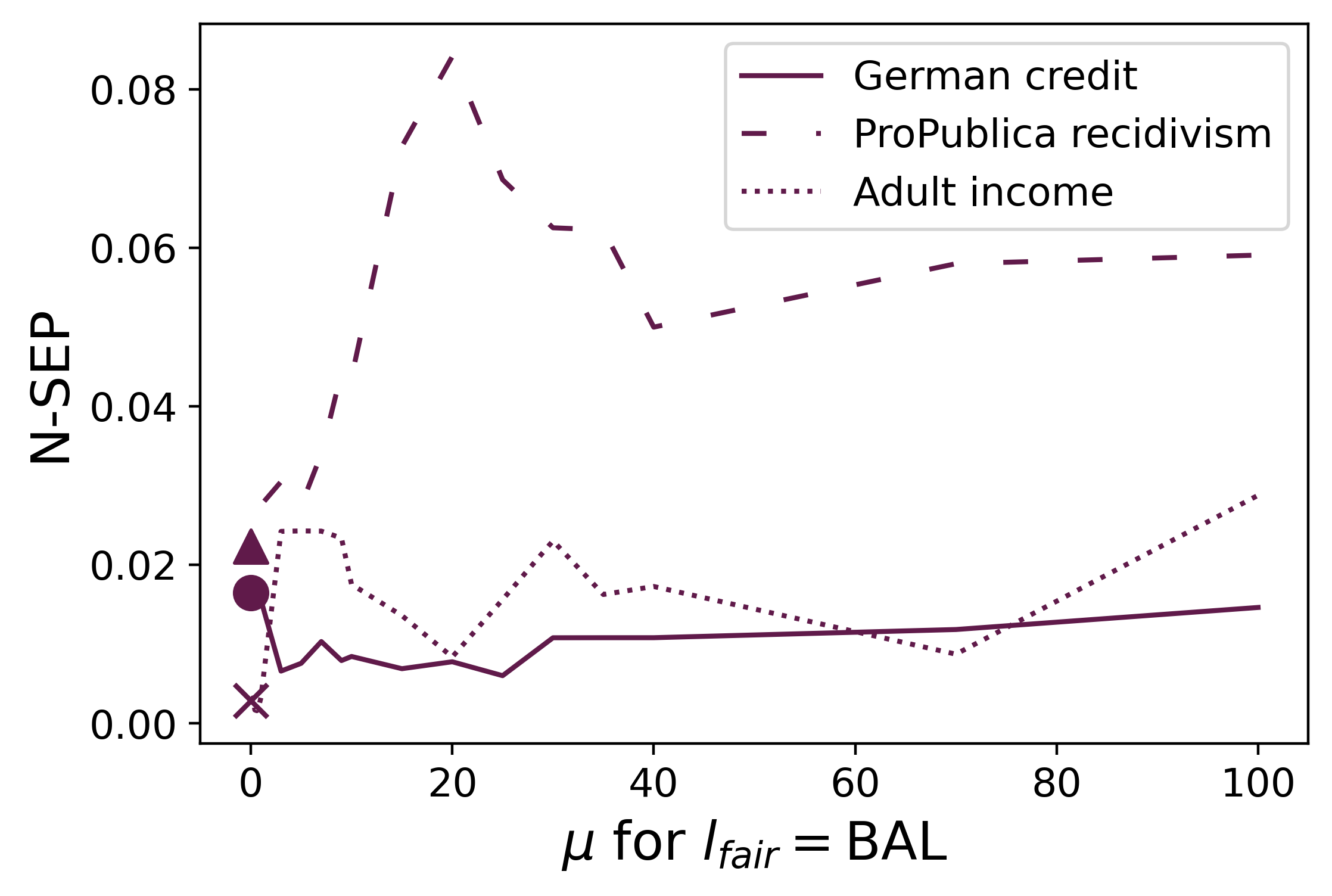}
         \caption{Balance $\rightarrow$ Separation}
         \label{fig:BAL--N-SEP}
     \end{subfigure}
     \hfill
     \begin{subfigure}[b]{0.23\textwidth}
         \centering
         \includegraphics[width=\textwidth]{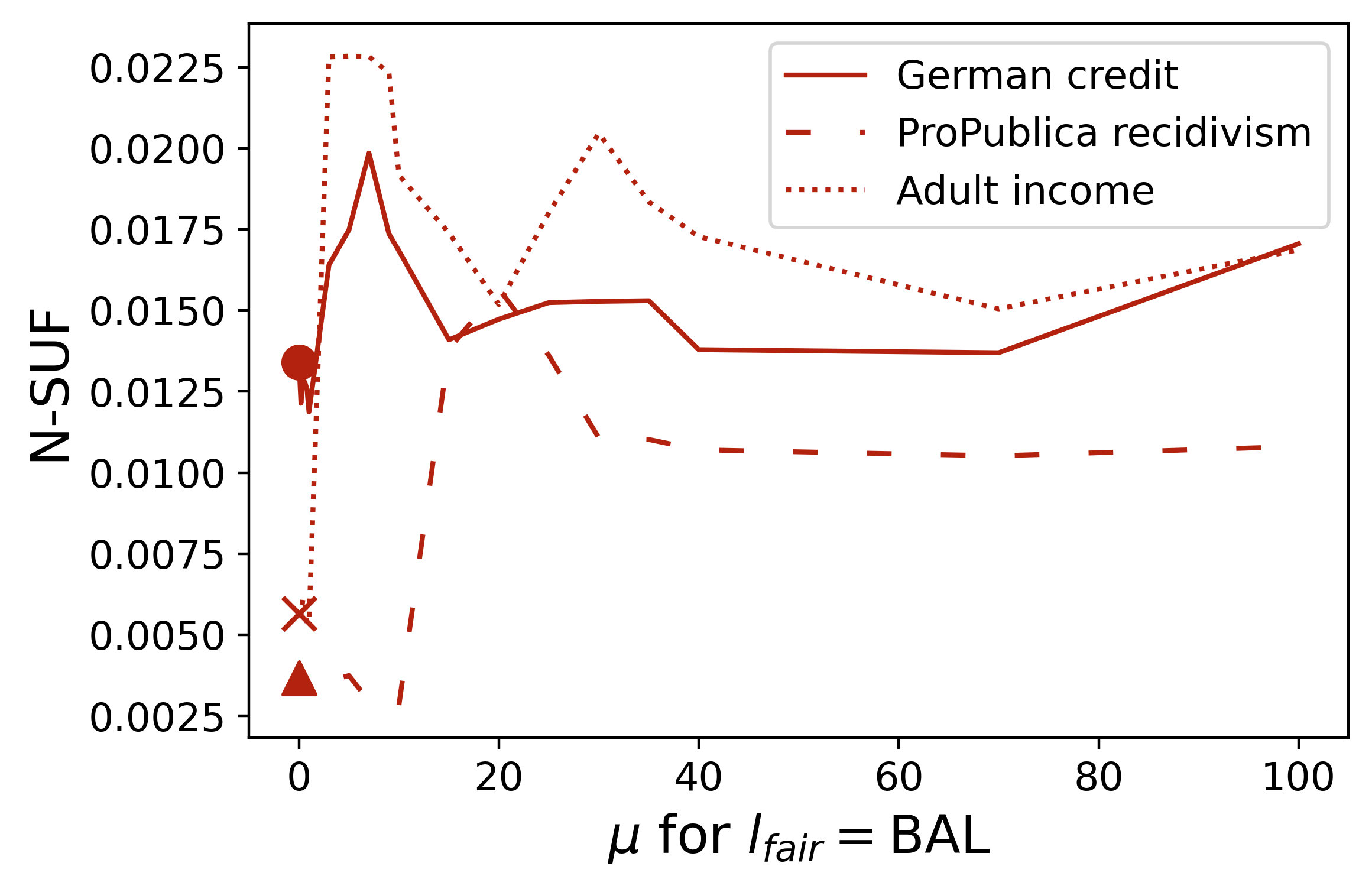}
         \caption{Balance $\rightarrow$ Sufficiency}
         \label{fig:BAL--N-SUF}
     \end{subfigure}
     \hfill
     \begin{subfigure}[b]{0.23\textwidth}
         \centering
         \includegraphics[width=\textwidth]{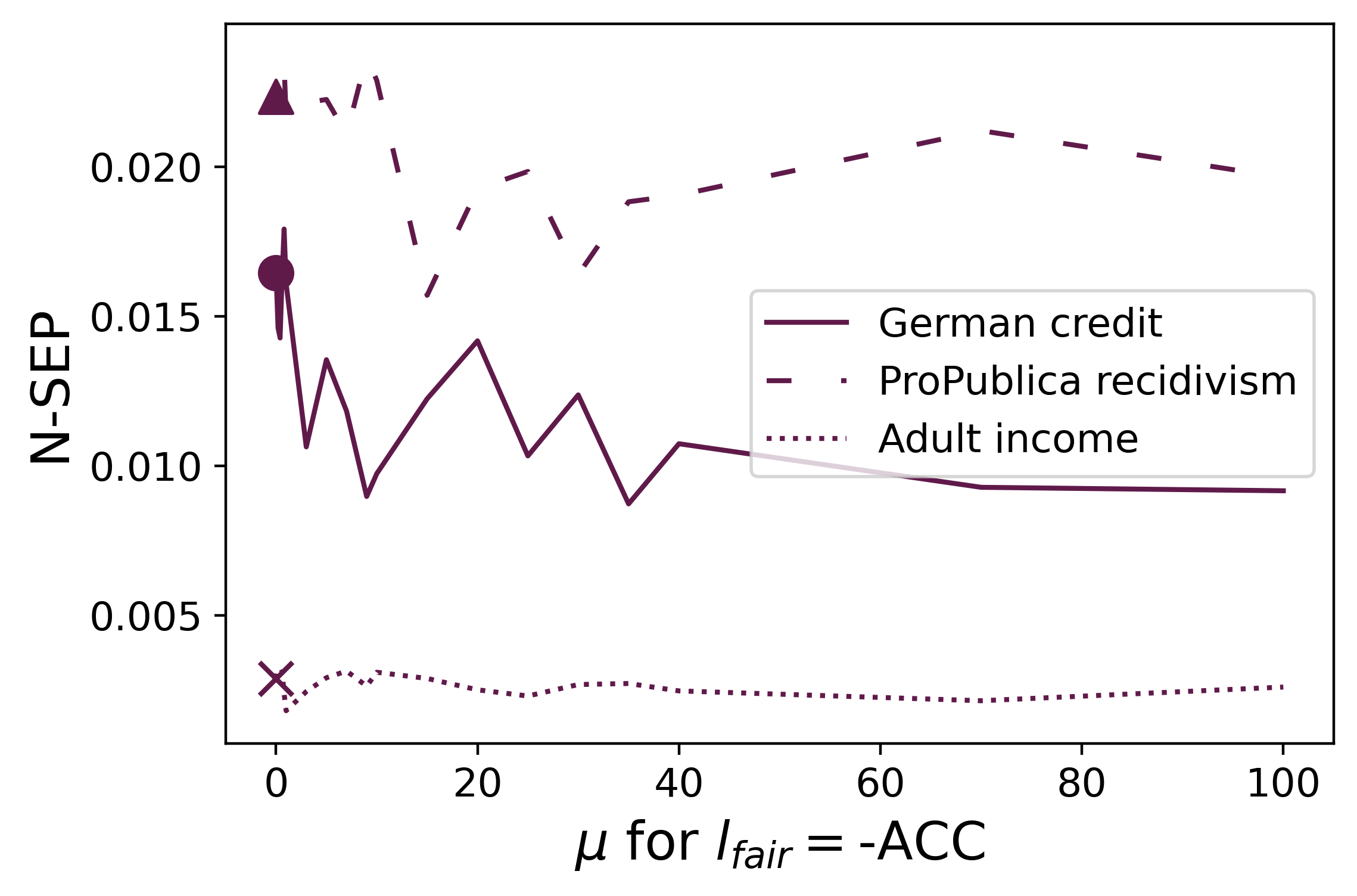}
         \caption{Accuracy $\rightarrow$ Separation}
         \label{fig:-ACC--N-SEP}
     \end{subfigure}
     \hfill
     \begin{subfigure}[b]{0.23\textwidth}
         \centering
         \includegraphics[width=\textwidth]{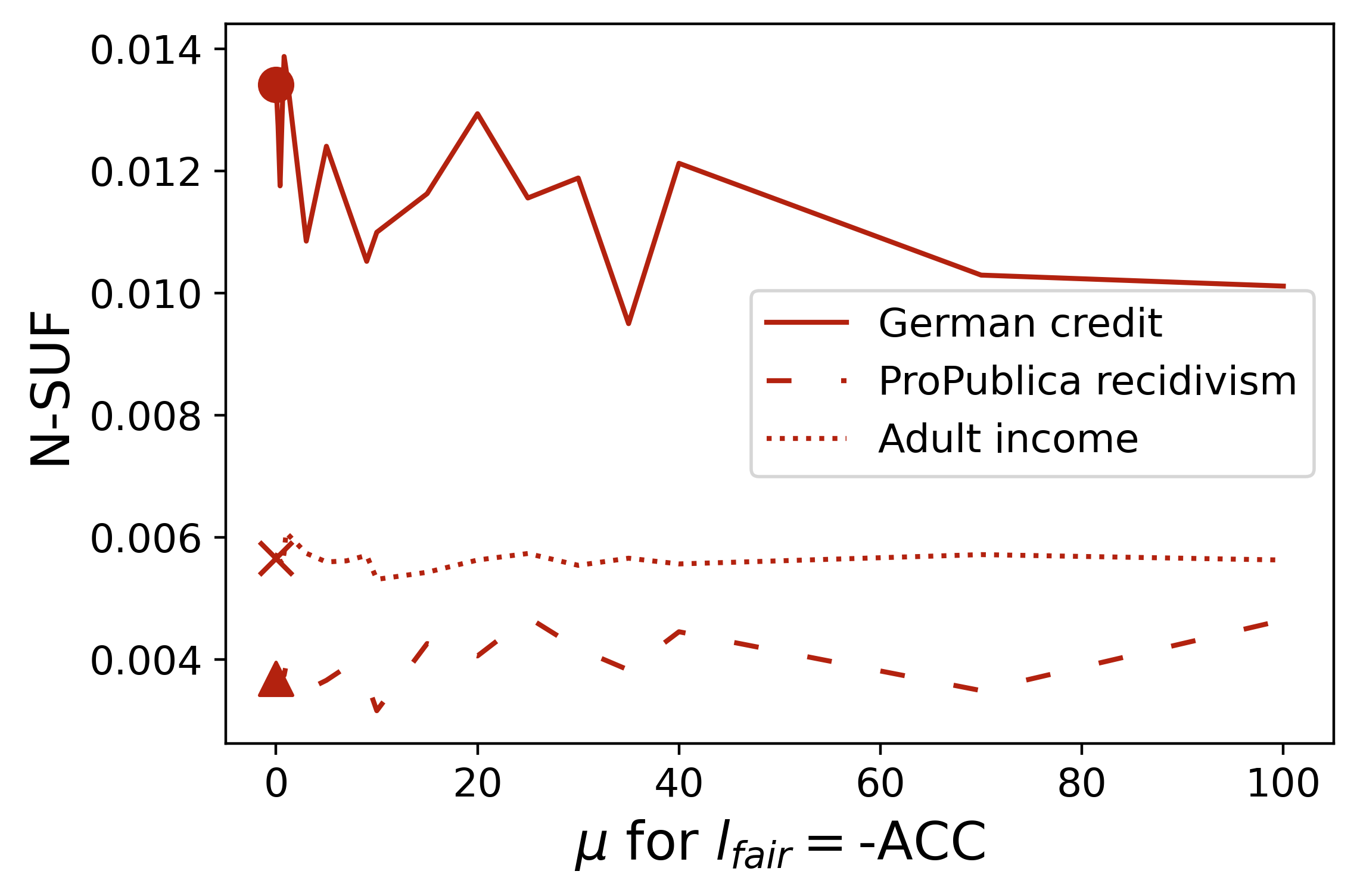}
         \caption{Accuracy $\rightarrow$ Sufficiency}
         \label{fig:-ACC--N-SUF}
     \end{subfigure}
        \caption{Indirect regularization of separation and sufficiency through negative accuracy and balance.}
        \label{fig:accuracy-balance}
\end{figure}

We examined the effects of using negative accuracy and balance as regularizers because they are part of the decomposition of both sufficiency and separation. We found that using -ACC and BAL did not yield stable improvements of either separation (N-SEP) or sufficiency (N-SUF) when used as fairness regularizers; results are in Figure \ref{fig:accuracy-balance}.

We investigated why this indirect regularization does not work. For this, we examined the effect of -ACC regularization on the components of separation and sufficiency (N-IND, BAL, -ACC) and found that the additional accuracy regularization does not substantially improve any of the components (see Figure \ref{fig:-ACC-effects} in the Appendix). As noted above, this may be due to the fact that optimizing the model for cross-entropy already indirectly enforces accuracy. Examining the effect of BAL regularization on the components of separation and sufficiency (N-IND, BAL, -ACC), we found that while balance itself improves, regularizing with balance also substantially decreases accuracy (see Figure \ref{fig:BAL-effects} in the Appendix). This may be the reason why the indirect regularization between sufficiency and separation does not work: enforcing balance seems to decrease accuracy too much.

\section{Conclusion}\label{sec:discussion}

In this paper, we have used information-theoretic formulations of fairness metrics to perform fairness regularization. We found, first, that using information-theoretic formulations of independence, sufficiency and separation in regularization works reasonably well. Second, there are indirect regularization relations between independence and separation: Enforcing independence improves separation and vice versa. This shows that it is possible to increase the degree to which fairness measures are satisfied, even though it is mathematically impossible to satisfy these measures perfectly at the same time. This finding is valuable for cases where stakeholders find both independence and separation to be appropriate fairness metrics. Third, we have not been able to verify indirect regularization between separation and sufficiency. We have traced this back to the balance term, which seems to counteract accuracy.


Future work could include the following issues: Formulations of degrees of fairness do not need to be based on information theory; it would be valuable to compare our results to those of other formulations of ``degrees of fairness'' to better understand the advantages and drawbacks of the information-theoretic formulation. In particular, it is possible to combine non-information-theoretic regularization with information-theoretic evaluation. Then, many conceptual problems are open: How should we interpret the tradeoffs we found in the present work? What meaning can we assign to degrees of fairness --- outside the extreme values? Finally, we have not fully explored the parameter space in that we have only explored linear scalings of terms in the decompositions of separation and sufficiency. For example, it is possible that separation and sufficiency can be improved at the same time by a judicious combination of weights for accuracy, balance, and independence.

\section*{Acknowledgments}

We thank the members of the NRP-77 project "Socially acceptable AI" for their useful feedback. We also appreciate the helpful feedback from the participants of the seminar "Scientific Writing II" at the University of Zurich. Thank you to Ryan Carey for suggesting the \textit{gap} terminology for measuring degrees of fairness. TR was supported by the Swiss National Science Foundation, grant number 197504. CH was supported by the National Research Programme “Digital Transformation” (NRP 77) of the Swiss National Science Foundation (SNSF), grant number 187473.

\bibliography{main}

\clearpage

\appendix

\section{Technical Appendix}

\subsection{Proofs of Propositions \ref{suff decomposition} and \ref{sep decomposition}}\label{app:proofs}

Both propositions can be proved by a repeated use of the chain rule and symmetry of (conditional) mutual information. The chain rule for conditional mutual information says that, for three random variables $X, Y, Z$, it holds $I(X;Y,Z) = I(X;Z) + I (X;Y \mid Z)$, symmetry says that $I(X;Y\mid Z) = I(Y;X\mid Z)$, and that $I(X;Y) = I(Y;X)$; see \citet{cover2006}.

\emph{Proof of Proposition \ref{suff decomposition}:} A repeated use of the chain rule and symmetry yields the decomposition of sufficiency: \begin{eqnarray*}
I(Y;A\mid R) &=& - I (Y;R) + I (R,A;Y) \\
&=& - I (Y;R) + I(Y;R \mid A) + I(A;Y). 
\end{eqnarray*} 
\begin{flushright}
$\square$
\end{flushright}

\emph{Proof of Proposition \ref{sep decomposition}:} By applying the same rules as for the decomposition of sufficiency, we get the decomposition of separation: \begin{eqnarray*}
I(R;A\mid Y) &=& - I (Y;R) + I(Y,A;R)\\
&=& - I (Y;R) + I(Y;R \mid A) + I(A;R).
\end{eqnarray*}
\begin{flushright}
$\square$
\end{flushright}

\subsection{Reproducibility checklist}\label{sec:reproducibility-checklist}

\paragraph{Code availability} The code can be found in the public repository \url{https://github.com/hcorinna/gradual-compatibility}.

\paragraph{Hardware} OS: Microsoft Windows 10 Enterprise, RAM: 16GB
\paragraph{Software} Python 3.7.9 with Anaconda 4.10.3. The packages are specified in the \texttt{requirements.txt} file in the code repository.

\paragraph{Evaluation runs} We ran the experiments with 5-fold cross-validation and a specified random seed.

\paragraph{Hyperparameters} scipy's minimize function: \texttt{maxiter}: 1000; logistic regression's fit: \texttt{initial weights}: 1 for every feature; k-fold cross-validations: \texttt{shuffle}: true

\paragraph{Running the experiments from the repository} Instructions for running the experiments can be found in the README file in the repository. We only provide a short overview here. First, the lambda values for the L2 regularization have to be calculated for each dataset. This can be done by running the Jupyter notebook \texttt{L2 - regularization.ipynb}. Next, the files \texttt{evaluate\_german.py}, \texttt{evaluate\_compas.py} and \texttt{evaluate\_adult.py} have to be run. These files will evaluate $\mu \in [0, 190]$ for all possible values of $l_{fair}$. The results of the evaluations will be saved in the folder \texttt{evaluations}. Finally, the notebook \texttt{Plots - direct and indirect effects.ipynb} can be run to create the plots shown in the paper. Configurations (color schemes, optimization method, values of $\mu$, etc.) can be changed in the \texttt{config.py} file. This is also the place to configure other datasets to test.

\newpage

\subsection{Additional figures}

\begin{figure}[ht] 
     \centering
     \begin{subfigure}[b]{0.42\textwidth}
         \centering
         \includegraphics[width=\textwidth]{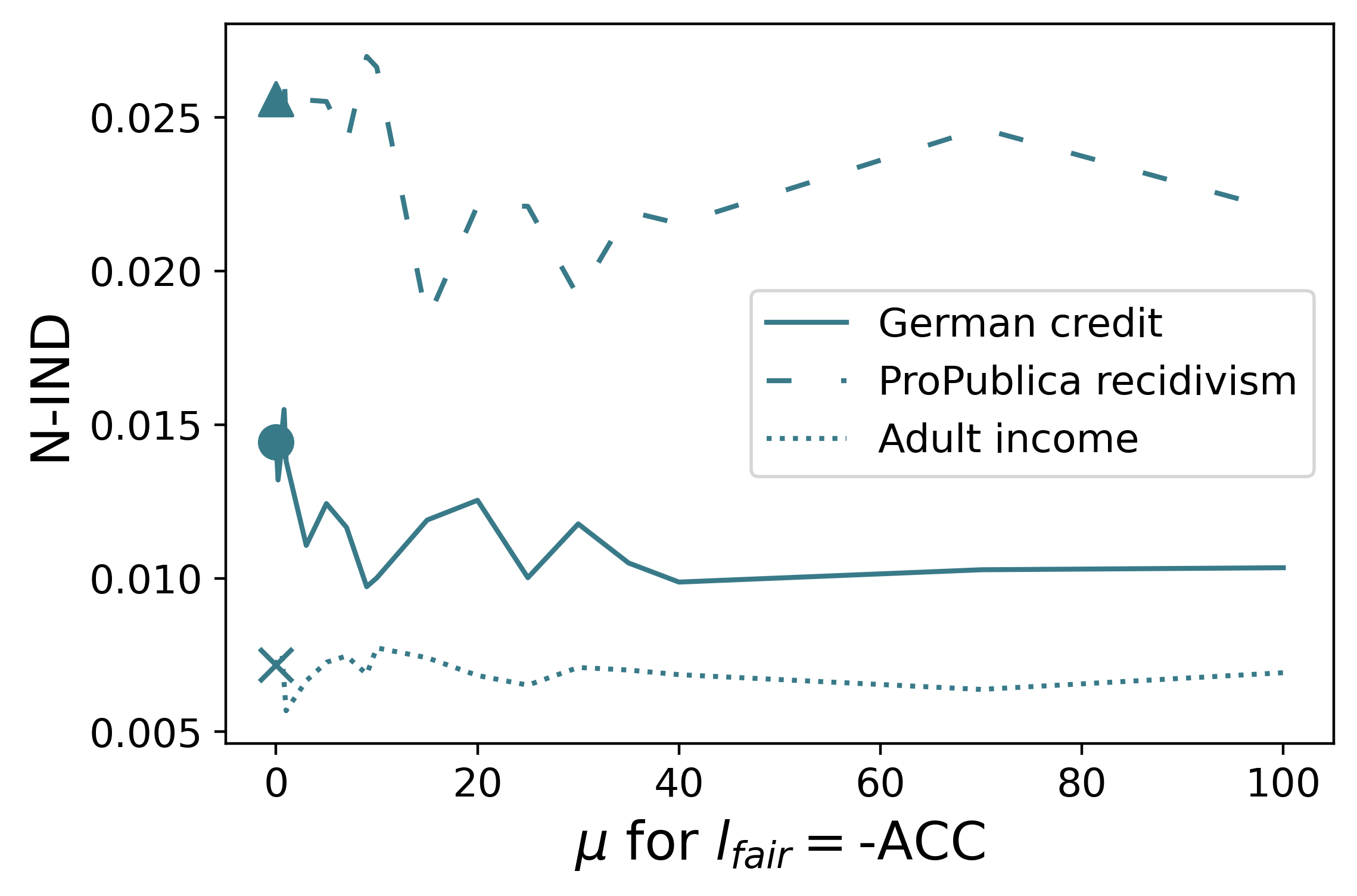}
         \caption{Accuracy $\rightarrow$ Independence}
     \end{subfigure}
     \hfill
     \begin{subfigure}[b]{0.42\textwidth}
         \centering
         \includegraphics[width=\textwidth]{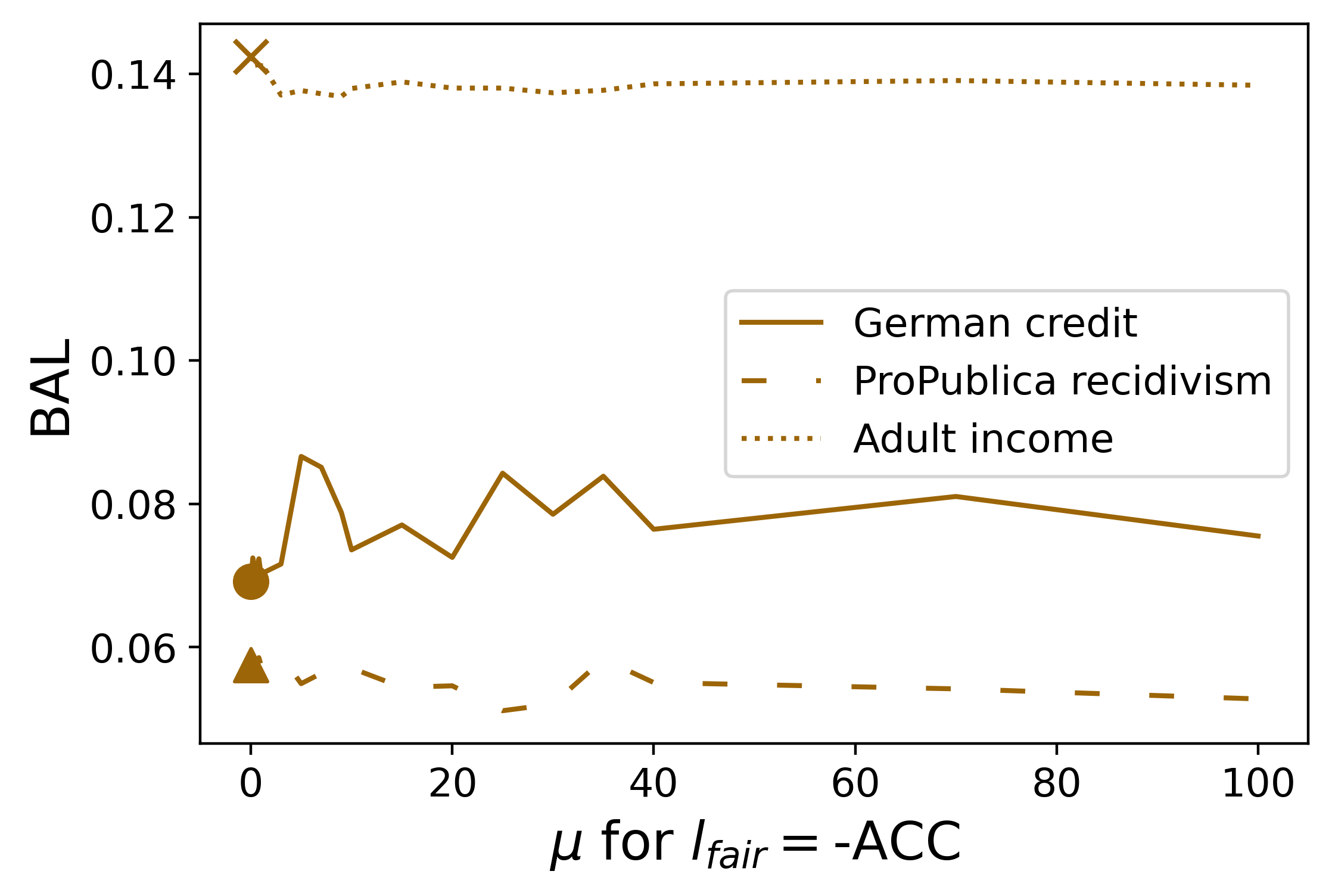}
         \caption{Accuracy $\rightarrow$ Balance}
     \end{subfigure}
     \hfill
     \begin{subfigure}[b]{0.42\textwidth}
         \centering
         \includegraphics[width=\textwidth]{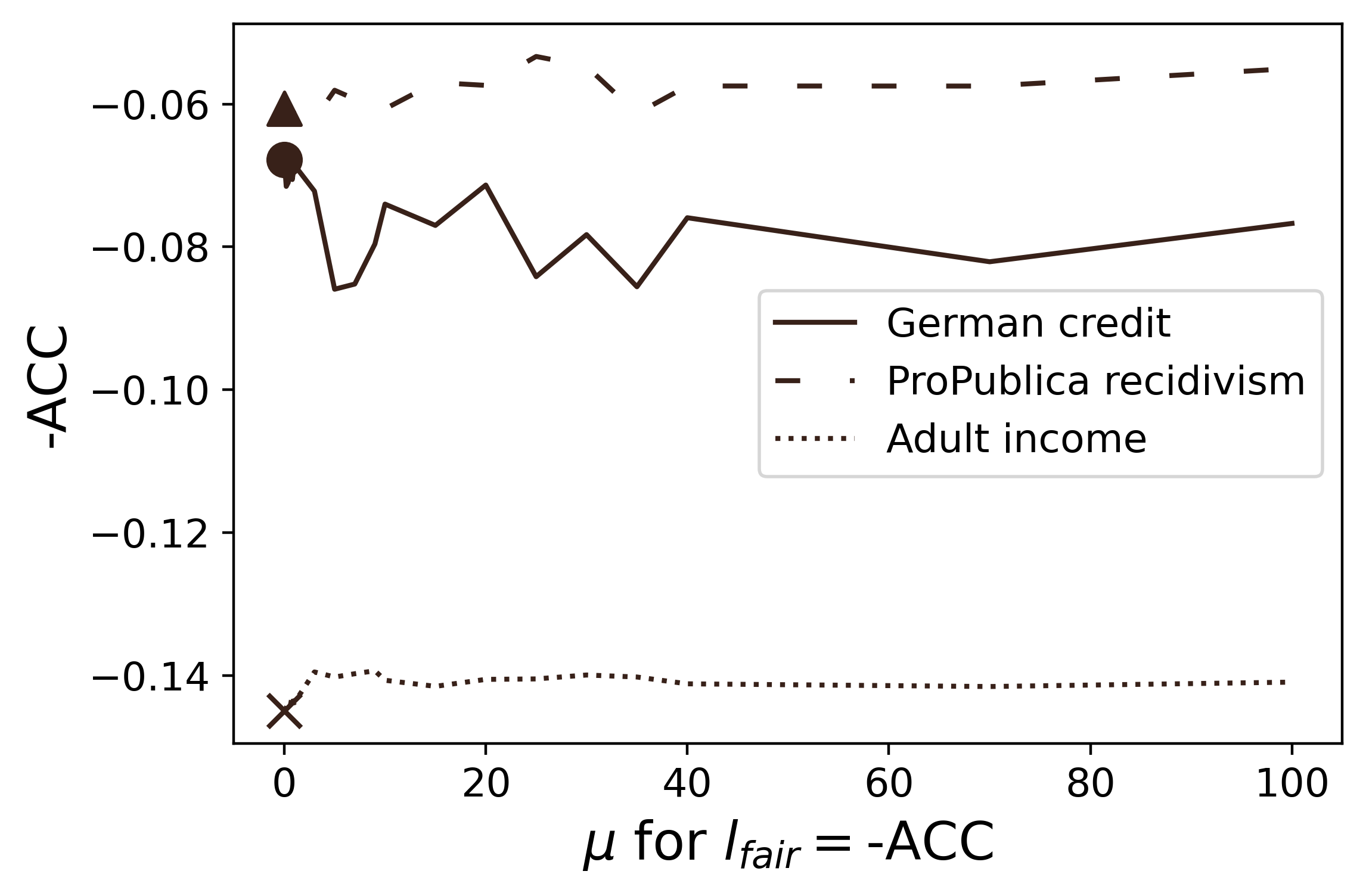}
         \caption{Accuracy $\rightarrow$ Accuracy}
     \end{subfigure}
        \caption{Effects of negative accuracy regularization on changeable components of separation and sufficiency.}
        \label{fig:-ACC-effects}
\end{figure}

\begin{figure}[ht] 
     \centering
     \begin{subfigure}[b]{0.42\textwidth}
         \centering
         \includegraphics[width=\textwidth]{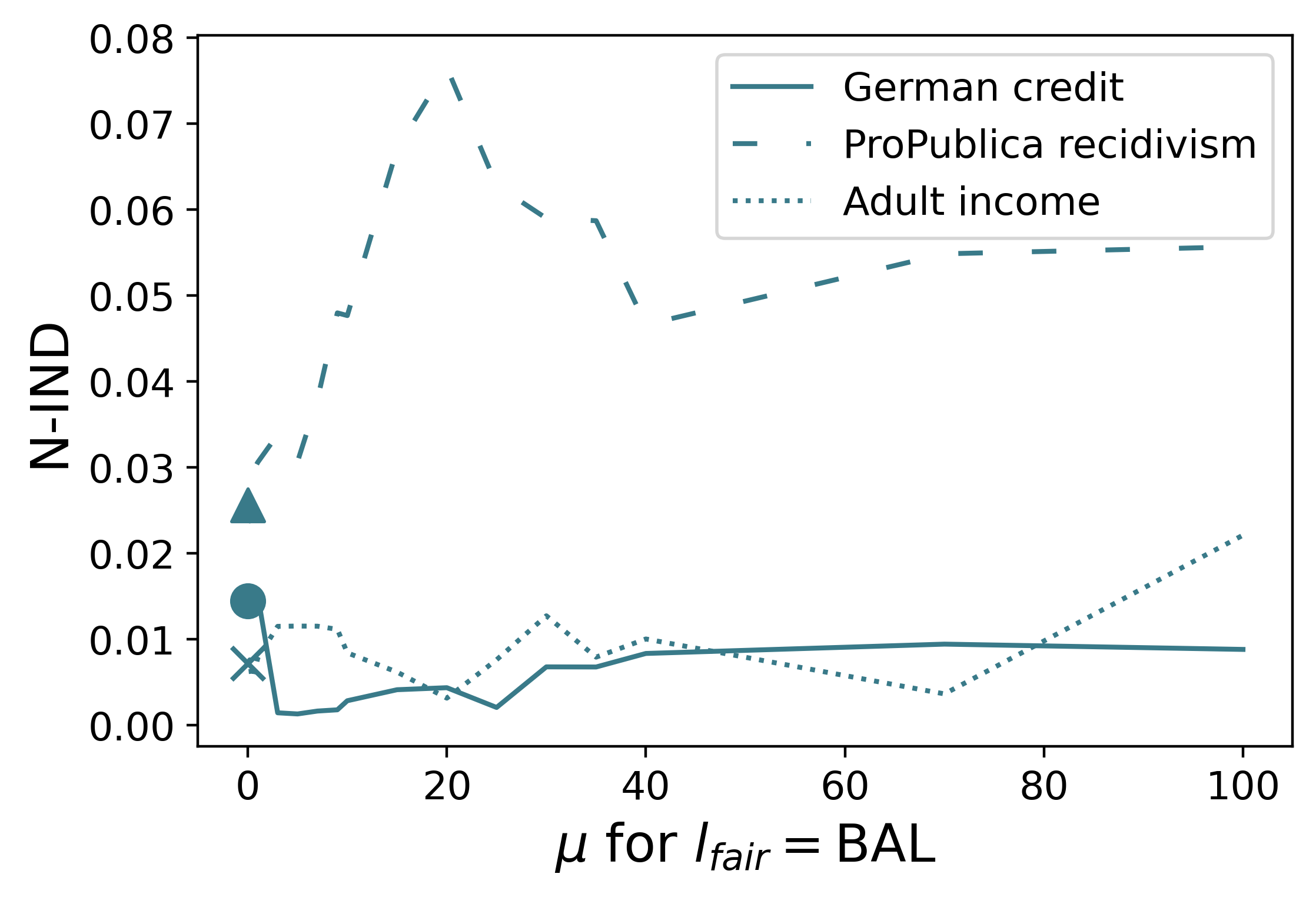}
         \caption{Balance $\rightarrow$ Independence}
     \end{subfigure}
     \hfill
     \begin{subfigure}[b]{0.42\textwidth}
         \centering
         \includegraphics[width=\textwidth]{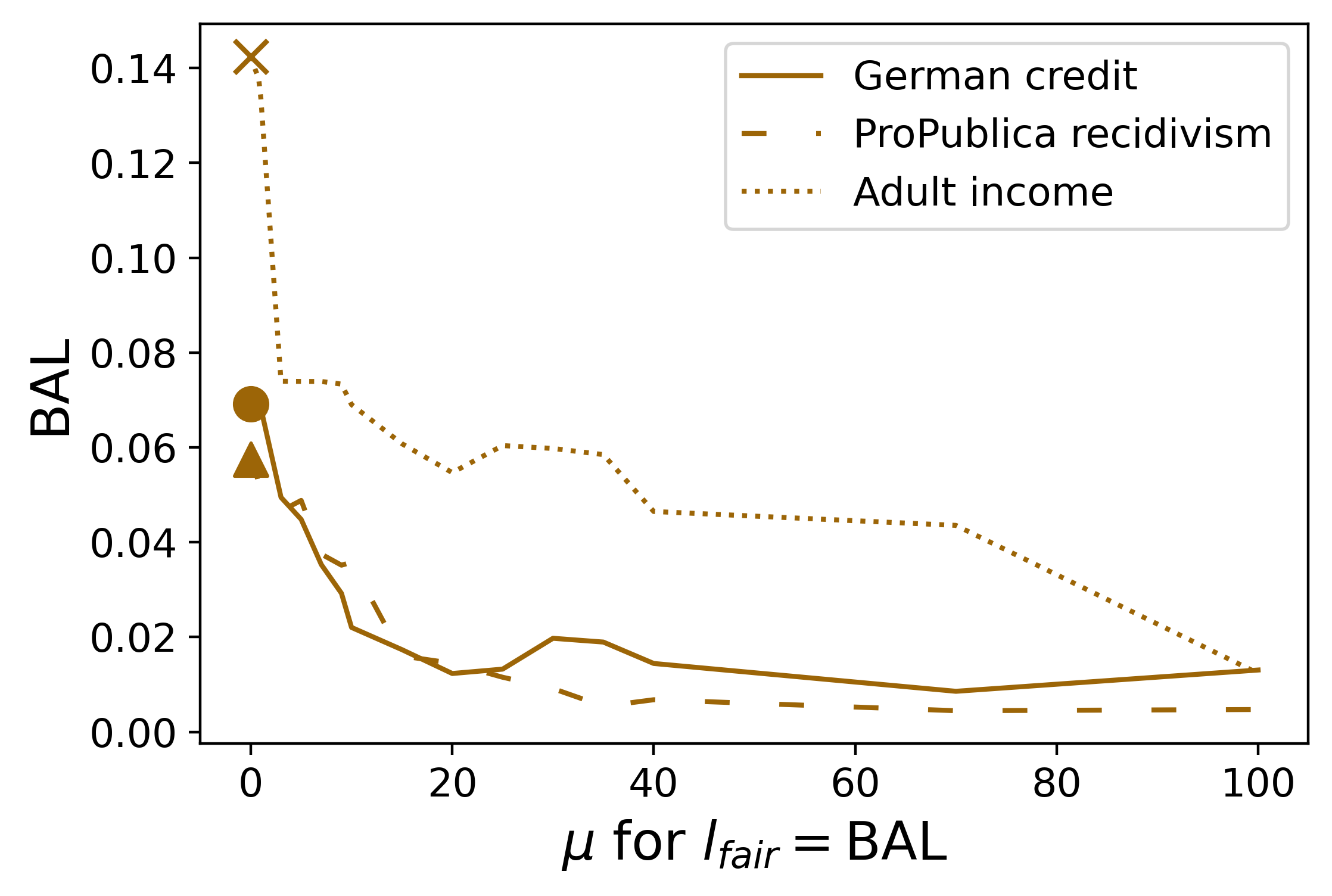}
         \caption{Balance $\rightarrow$ Balance}
     \end{subfigure}
     \hfill
     \begin{subfigure}[b]{0.42\textwidth}
         \centering
         \includegraphics[width=\textwidth]{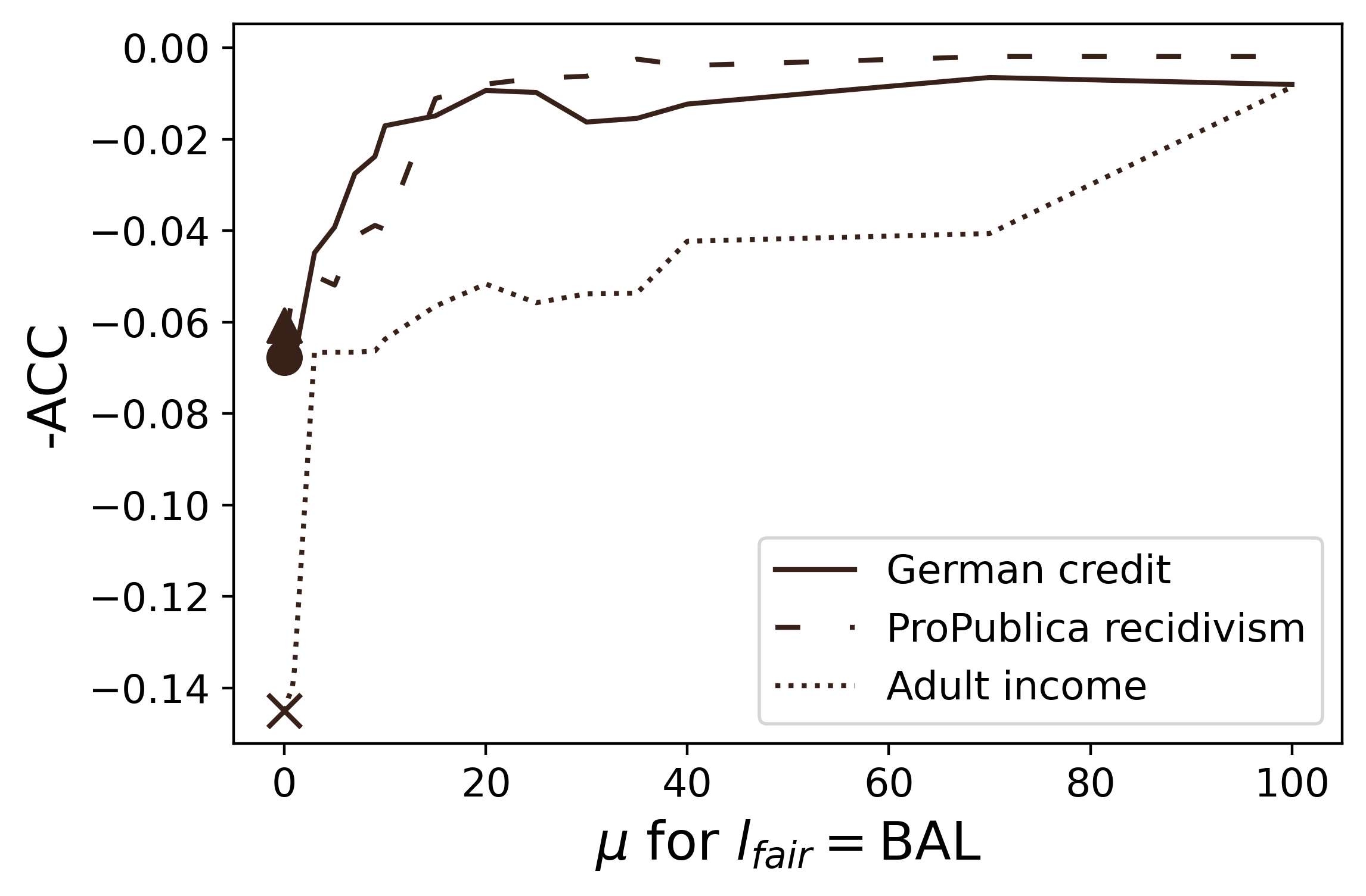}
         \caption{Balance $\rightarrow$ Accuracy}
     \end{subfigure}
        \caption{Effects of balance regularization on changeable components of separation and sufficiency.}
        \label{fig:BAL-effects}
\end{figure}

\begin{figure}[ht] 
     \centering
     \begin{subfigure}[b]{0.42\textwidth}
         \centering
         \includegraphics[width=\textwidth]{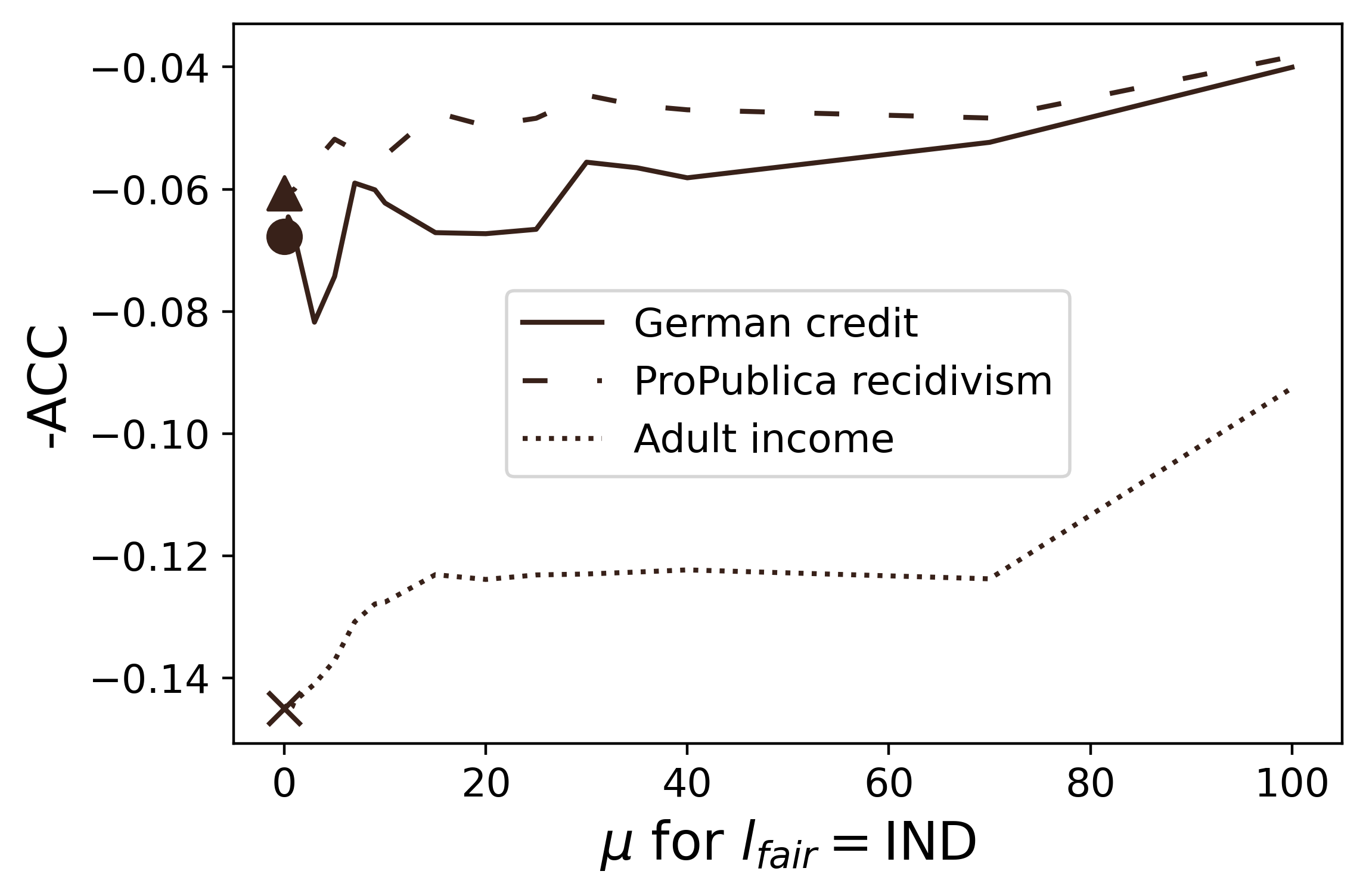}
         \caption{Independence $\rightarrow$ Accuracy}
     \end{subfigure}
     \hfill
     \begin{subfigure}[b]{0.42\textwidth}
         \centering
         \includegraphics[width=\textwidth]{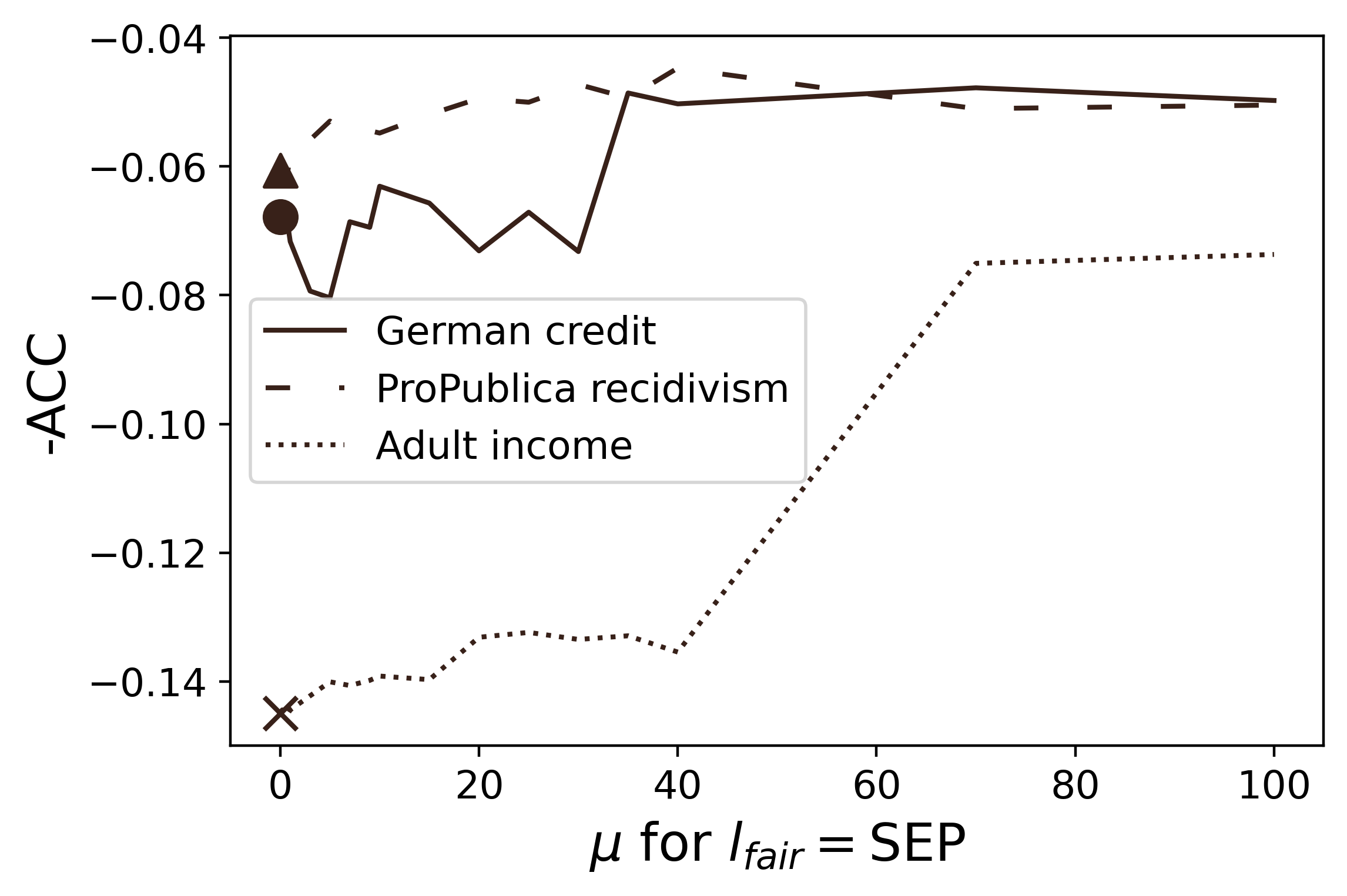}
         \caption{Separation $\rightarrow$ Accuracy}
     \end{subfigure}
     \hfill
     \begin{subfigure}[b]{0.42\textwidth}
         \centering
         \includegraphics[width=\textwidth]{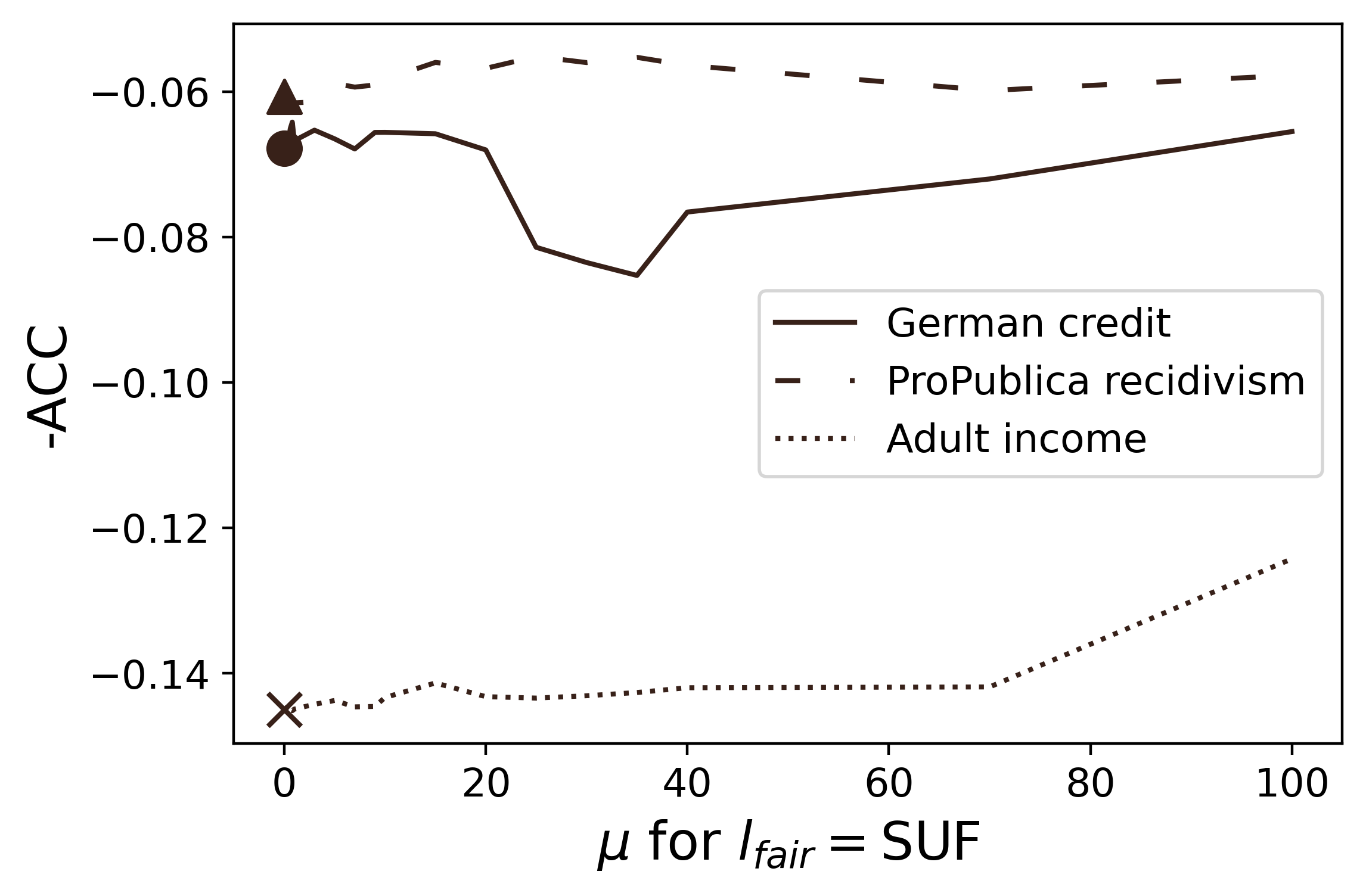}
         \caption{Sufficiency $\rightarrow$ Accuracy}
     \end{subfigure}
        \caption{Effects of independence, separation and sufficiency regularization on accuracy.}
        \label{fig:effects-on-ACC}
\end{figure}

\end{document}